\documentclass{article}

\usepackage{microtype}
\usepackage{graphicx}
\usepackage{subcaption}
\usepackage{booktabs} 
\usepackage{xcolor} 
\usepackage{hyperref}

\usepackage[accepted]{icml2026}
\usepackage{amsmath}
\usepackage{amssymb}
\usepackage{mathtools}
\usepackage{amsthm}

\usepackage{float}
\usepackage[capitalize,noabbrev]{cleveref}

\theoremstyle{plain}

\theoremstyle{definition}

\theoremstyle{remark}

\usepackage{booktabs} 
\usepackage{multirow} 
\usepackage{array}    
\usepackage{comment}
\usepackage{ulem}
\usepackage{float}

\usepackage[textsize=tiny]{todonotes}

\icmltitlerunning{Efficient Video Transfer for Vision-Language-Action Data Augmentation}

\definecolor{phasecolor}{RGB}{0, 70, 140} 
\definecolor{darkgreen}{rgb}{0,0.7,0}
\newcommand{\blue}[1]{{\color{blue}{#1}}}
\newcommand{\red}[1]{{\color{red}{#1}}}
\newcommand{\green}[1]{{\color{darkgreen}{#1}}}
\definecolor{mygraytext}{gray}{.5}

\newcommand{\inc}[1]{\scriptsize{\green{(+#1)}}}
\newcommand{\dec}[1]{\scriptsize{\red{(-#1)}}}

\usepackage{tcolorbox}
\tcbuselibrary{breakable}
\newtcolorbox{promptbox}{
  breakable,
  colback=gray!5,
  colframe=gray!40,
  boxrule=0.5pt,
  arc=2pt,
  left=6pt,
  right=6pt,
  top=6pt,
  bottom=6pt,
  fontupper=\small\ttfamily
}

\begin{document}

\twocolumn[
  \icmltitle{Seeing Realism from Simulation: Efficient Video Transfer for Vision-Language-Action Data Augmentation}



  \icmlsetsymbol{equal}{*}
  \icmlsetsymbol{corr}{$\dagger$}
  \begin{icmlauthorlist}
    \icmlauthor{Chenyu Hui}{sjtu,xjtu,equal}
    \icmlauthor{Xiaodi Huang}{sjtu,cas,equal}
    \icmlauthor{Siyu Xu}{sydney}
    \icmlauthor{Yunke Wang}{sydney}
    \icmlauthor{Shan You}{comp}
    \icmlauthor{Fei Wang}{ustc}
    \icmlauthor{Tao Huang}{sjtu,corr}
    \icmlauthor{Chang Xu}{sydney}
  \end{icmlauthorlist}

  \icmlaffiliation{xjtu}{Xi'an Jiaotong University}
  \icmlaffiliation{cas}{Institute of Automation, Chinese Academy of Sciences}
  \icmlaffiliation{sydney}{The University of Sydney}
  \icmlaffiliation{ustc}{University of Science and Technology of China}
  \icmlaffiliation{comp}{SenseTime Research}
  \icmlaffiliation{sjtu}{Shanghai Jiao Tong University}

  \icmlcorrespondingauthor{Tao Huang}{t.huang@sjtu.edu.cn}

  \icmlkeywords{Machine Learning, ICML}

  \vskip 0.3in
]



\printAffiliationsAndNotice{$^*$Equal contribution. Work was done during internship at SJTU.}

\begin{abstract}

Vision-language-action (VLA) models typically rely on large-scale real-world videos, whereas simulated data, despite being inexpensive and highly parallelizable to collect, often suffers from a substantial visual domain gap and limited environmental diversity, resulting in weak real-world generalization. We present an efficient  video augmentation framework that converts simulated VLA videos into realistic training videos while preserving task semantics and action trajectories. Our pipeline extracts structured conditions from simulation via video semantic segmentation and video captioning, rewrites captions to diversify environments, and uses a conditional video transfer model to synthesize realistic videos. To make augmentation practical at scale, we introduce a diffusion feature-reuse mechanism that reuses video tokens across adjacent timesteps to accelerate generation, and a coreset sampling strategy that identifies a compact, non-redundant subset for augmentation under limited computation.
Extensive experiments on Robotwin 2.0, LIBERO, LIBERO-Plus, and a real robotic platform demonstrate consistent improvements.
For example, our method improves RDT-1B by 8\% on Robotwin 2.0, and boosts $\pi_0$ by 5.1\% on the more challenging LIBERO-Plus benchmark. Code is available at: \href{https://github.com/nanfangxiansheng/Seeing-Realism-from-Simulation}{CODE}.

\end{abstract}

\section{Introduction}
The advent of large-scale Vision-Language-Action (VLA) models has marked a significant milestone in robotics~\cite{liu2024rdt,black2410pi0,intelligence2504pi0,intelligence2025pi06vlalearnsexperience,liu2024rdt,zhao2023learning,xu2025vla}, enabling robots to interpret natural language instructions and execute complex manipulation tasks. These models typically rely on extensive datasets of real-world robotic trajectories to learn generalizable policies. However, the collection of such real-world data is inherently costly, time-consuming, and difficult to scale, presenting a major bottleneck for broader development and deployment. In contrast, simulated data offers a highly inexpensive alternative. But as noted in recent studies, models that achieve near-perfect performance within a specific simulation benchmark often fail catastrophically when faced with minor perturbations in object layout, lighting, camera viewpoint, or instruction phrasing in scenarios with disturbances~\cite{fei2025libero,zhou2025libero,xu2025affordance,pei2025action}, revealing that many models merely memorize training sequences rather than learning robust, semantic task understanding.

\begin{figure*}[htbp]
    \centering
    \includegraphics[width=1\linewidth,height=0.43\linewidth]{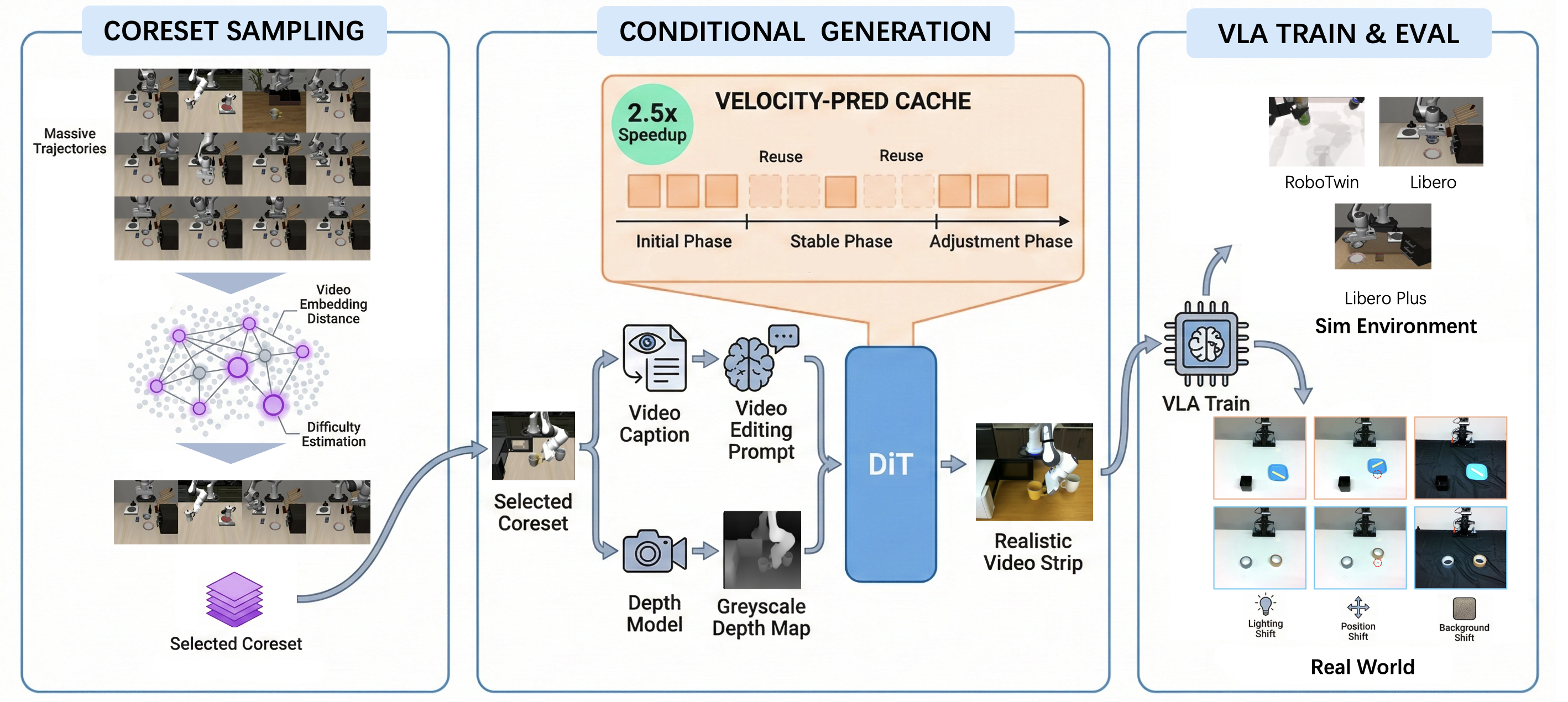}
        \caption{Overall framework of the proposed method. Given a large-scale simulation training set, we introduce a coreset sampling algorithm to select important and diverse samples, which are then augmented to realistic video strips and used for training.
    }
    \label{fig:overall_framework}
\end{figure*}

We address the critical lack of visual and environmental diversity in robotic training data by proposing an efficient, scalable video augmentation framework. Our approach transforms source video sequences into visually diverse training videos while strictly preserving underlying task semantics and action trajectories. The pipeline first extracts structured conditions from simulation, diversifies the environmental context, and then synthesizes high-fidelity videos using a conditional video diffusion model. To mitigate the substantial computational cost of video transfer, we propose accelerating the diffusion generation by caching velocity  during the denoising process. 
We design adaptive strategies that optimize the trade-off between quality and efficiency. Consequently, this velocity caching mechanism achieves over a 60\% reduction in generation time while maintaining high model accuracy.

Furthermore, a coreset sampling method is designed to achieve the efficient scalability by only augmenting a subset in the large-scale dataset. For the sake of sampling diverse trajectories that maximize the performance, we build a graph structure that balances the difficulty and diversity among data nodes, where the difficulty and diversity are estimated by a video embedding model and a pretrained VLA policy. 
By prioritizing samples that are both challenging and highly-diverse, we ensure the significance of the augmentation effect in large-scale datasets.

We conduct extensive experiments across multiple benchmarks, including Robotwin 2.0~\cite{chen2025robotwin}, LIBERO~\cite{liu2023libero}, and the more challenging LIBERO-Plus~\cite{fei2025libero} to demonstrate the effectiveness of our framework. We also conduct real-world robotic experiments.

In summary, our contributions are threefold:
\begin{itemize}
    \item We present an end-to-end video transfer named efficient transfer framework that effectively bridges the sim-to-real gap for VLA data augmentation.
    \item We design a segmented (three-stage) velocity caching strategy, which better matches the dynamics of conditional video generation. It is proved that adopting this strategy can save over 60\% transferring time.
    \item We  propose a trajectory-level coreset formulation that combines policy difficulty (measured by the policy loss of RDT-1B), and visual diversity (measured via Cosmos-Embed1 representations), leading to more effective selection of high-value training trajectories.
\end{itemize}

\section{Related Work}

\subsection{Vision-Language-Action Models}
Since the development of VLM and its been  pivotal in advancing robotic control by providing
rich multi-modal representations. Recent VLA models are finetuned on large scale robotic trajectory data in an end to end mode to obtain more generalizable  policies~\cite{brohan2022rt,ebert2021bridge}. Widely used OpenVLA~\cite{kim2024openvla} is based on prismatic~\cite{karamcheti2024prismatic}. RDT combines the advantage of diffusion model and transformer~\cite{liu2024rdt}. ACT~\cite{zhao2023learning} introduced chunking for actions based on transformer. The OpenPi series like $\pi_0$~\cite{black2410pi0}, $\pi_{0.5}$~\cite{intelligence2504pi0} and the newest $\pi_{0.6}$~\cite{intelligence2025pi06vlalearnsexperience} gains significant performance superiority by enhancing existing VLMs with a dual-system VLA architecture, retaining the text-generation ability of the VLM, and combining with reinforcement learning, respectively.

\subsection{Video Generation}

Recent advances in diffusion-based video generation and style transfer have established this as a pivotal area in multimodal modeling, building upon foundational works like Stable Diffusion ~\cite{rombach2022high}, Stable Video Diffusion ~\cite{blattmann2023stable}. Contemporary research has expanded these frontiers: Sora ~\cite{brooks2024video} incorporates a temporal VAE and Transformer backbone; Wan ~\cite{wan2025wan} and VACE ~\cite{jiang2025vace} employ mixture-of-experts architectures; while NVIDIA's Cosmos series ~\cite{alhaija2025cosmos}, particularly Cosmos Transfer ~\cite{ali2025world}, provides a robust foundation for world model-based video transfer. Concurrently, generative techniques are increasingly applied to embodied data, with methods like Rebot ~\cite{fang2025rebot}, EgoDemoGen ~\cite{xu2025egodemogen}, EMMA ~\cite{dong2025emma}, Embodied Dreamer ~\cite{wang2025embodiedreamer}, Gigaworld-0 ~\cite{team2025gigaworld}, RoboTransfer~\cite{liu2025robotransfer} and Gigabrain-0 ~\cite{team2025gigabrain} demonstrating effective real-to-sim and sim-to-real data generation.

\subsection{Data Centric AI and Efficient Training}

Data-centric approaches have proven crucial for enhancing model performance and training efficiency, evolving from early heuristic-based filtering methods like CCNet~\cite{wenzek2020ccnet} and C4 ~\cite{raffel2020exploring} to sophisticated techniques emphasizing deduplication~\cite{lee2022deduplicating} and semantic redundancy elimination~\cite{abbas2023semdedup}. Beyond quality filtering, data diversity plays an equally vital role in robustness, as demonstrated by The Pile's emphasis on heterogeneous data~\cite{gao2020pile} and Falcon Series' success with web-scale curation~\cite{penedo2023refinedweb}. Domain-specific selection methods like DSIR~\cite{xie2023data} leverage importance resampling for target distribution alignment, while theoretical advances~\cite{sorscher2022beyond} show intelligent pruning can achieve exponential efficiency gains. This paradigm culminates in  $\mathbb{D}^2$ Pruning~\cite{maharana2023d2}, which advocates balanced coreset selection based on complementary diversity-difficulty criteria.
\section{Method}
We propose a VLA data augmentation framework to improve sim-to-real generalization. We first analyze the generalization failures of simulation-trained models, then introduce an end-to-end pipeline that generates realistic videos from semantic and structural cues. To scale augmentation efficiently, we develop a diffusion acceleration method based on velocity caching and a coreset sampling strategy for selective data generation.

\subsection{Motivation: The Brittleness of VLA Models}

VLA models trained on robotic manipulation data often achieve high in-distribution performance but fail under distribution shifts. The core issue is \textbf{limited visual and environmental diversity} in training data: simulated environments are overly clean and static, while real-world datasets cover only narrow conditions due to collection constraints. This lets models \textbf{overfit to spurious regularities} rather than learning robust instruction-to-action mappings.

\begin{figure}[t]
    \centering
    \includegraphics[width=\linewidth]{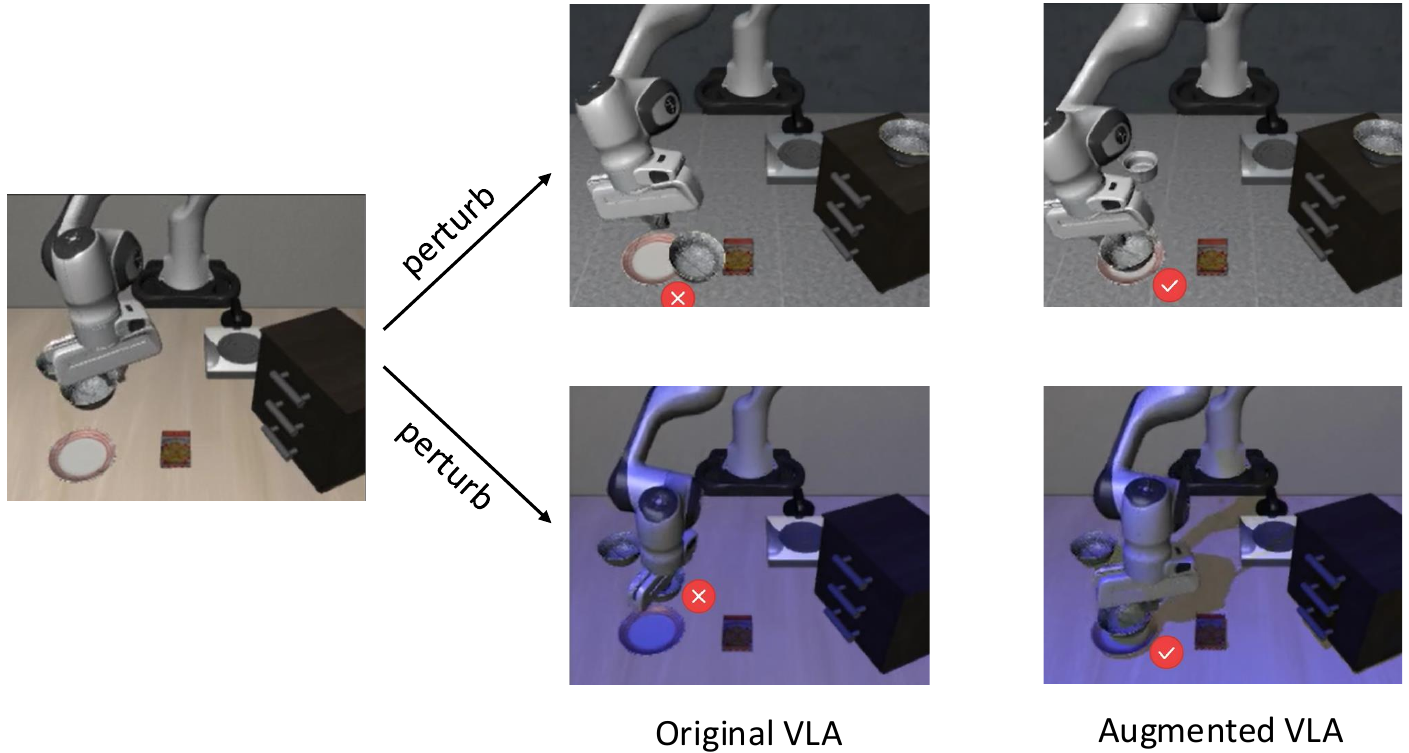}
    \vspace{-4mm}
    \caption{
    Examples from LIBERO-Plus evaluation. The baseline VLA model fails under environment perturbations such as texture change (upper) and lighting change (lower), while the model trained with our augmented data performs the tasks correctly, showing stronger generalization.
    }
    \label{fig:libero_cmp}
    \vspace{-4mm}
\end{figure}

Recent benchmarks quantify this brittleness: LIBERO-Plus~\cite{fei2025libero} shows success rates dropping from 95\% to under 30\% with minor perturbations in object layout, camera angle, or lighting; LIBERO-PRO~\cite{zhou2025libero} reports near-zero accuracy when object positions and instructions are modified. As illustrated in Figure~\ref{fig:libero_cmp}, baseline VLA models fail under such perturbations, revealing memorization of fixed action sequences rather than semantic task understanding.

The root cause lies in sterile training data that lacks the complexity of real-world contexts. Without diverse backgrounds, lighting, viewpoints, and object configurations, models latch onto superficial correlations~\cite{xu2025affordance,li2025vlas,fang2025intention} and under-utilize language inputs~\cite{fei2025libero,zhou2025libero}. This motivates our approach: augmenting training distributions with visually varied yet semantically faithful data to enable robust generalization.

\subsection{Diversifying Data via Conditional Video Transfer}
We present a video augmentation framework that transforms source robotic videos into visually diverse counterparts while preserving task semantics and action fidelity. As illustrated in Figure~\ref{fig:overall_framework}, our pipeline consists of four stages: caption generation, caption rewriting, structured condition extraction, and conditional video synthesis.

\noindent \textbf{Semantic and Structural Condition Extraction.}
We first extract descriptive captions from source videos using a temporal video captioning model (\textit{i.e.}, VideoChat2~\cite{li2023videochat}), which summarizes interactions, objects, and spatial relations. These captions are then rewritten by a large language model (\textit{i.e.}, Qwen3-8B~\cite{yang2025qwen3}) to introduce environmental variations such as background and object color changes, yielding diverse scene contexts while preserving task intent (details in \ref{sec:appendix_llm}). To maintain geometric consistency, we also extract depth maps from source videos as structural control signals, which provide stable, geometry-preserving guidance compared to alternatives like blur, edge, or segmentation~\cite{ravi2024sam}.

\noindent \textbf{Conditional Video Synthesis.}
The final stage employs a conditional video diffusion model (\textit{i.e.}, Cosmos-Transfer 2.5~\cite{ali2025world}) to synthesize realistic videos conditioned on the rewritten caption and depth input. The model iteratively denoises video tokens, generating temporally coherent scenes that retain the original action trajectory while varying the visual context. This process enriches the training distribution with diverse visual conditions without requiring additional real-world data collection.

\subsection{Efficient Video Generation via Velocity Caching}

Recent conditional video diffusion models such as Cosmos-Transfer~\cite{ali2025world} and Wan~\cite{wan2025wan} achieve strong visual fidelity but suffer from prohibitively high inference costs\footnote{Cosmos-Transfer 1 takes $\sim$40 minutes for a 5-second 720p video on one A100 GPU.}, posing a major obstacle to large-scale data augmentation. We propose a velocity caching mechanism that exploits temporal redundancy in the diffusion process to significantly accelerate generation while preserving quality.

\noindent \textbf{Observation: Redundancy in Denoising.}
In flow-based diffusion models~\cite{lipman2022flow}, each denoising step computes a velocity field $v_\theta(x_t, t)$ to update the latent:
\begin{equation}
x_{t+1} = x_t + \Delta t \cdot v_\theta(x_t, t).
\end{equation}
This velocity prediction accounts for over 70\% of per-step runtime due to the full forward pass through a large video diffusion transformer. Inspired by recent caching strategies~\cite{liu2025timestep,zhou2025less,li2023diffnas}, we analyze the temporal dynamics of velocity predictions during denoising. 

\begin{figure}[t]
\centering
\includegraphics[width=1\linewidth]{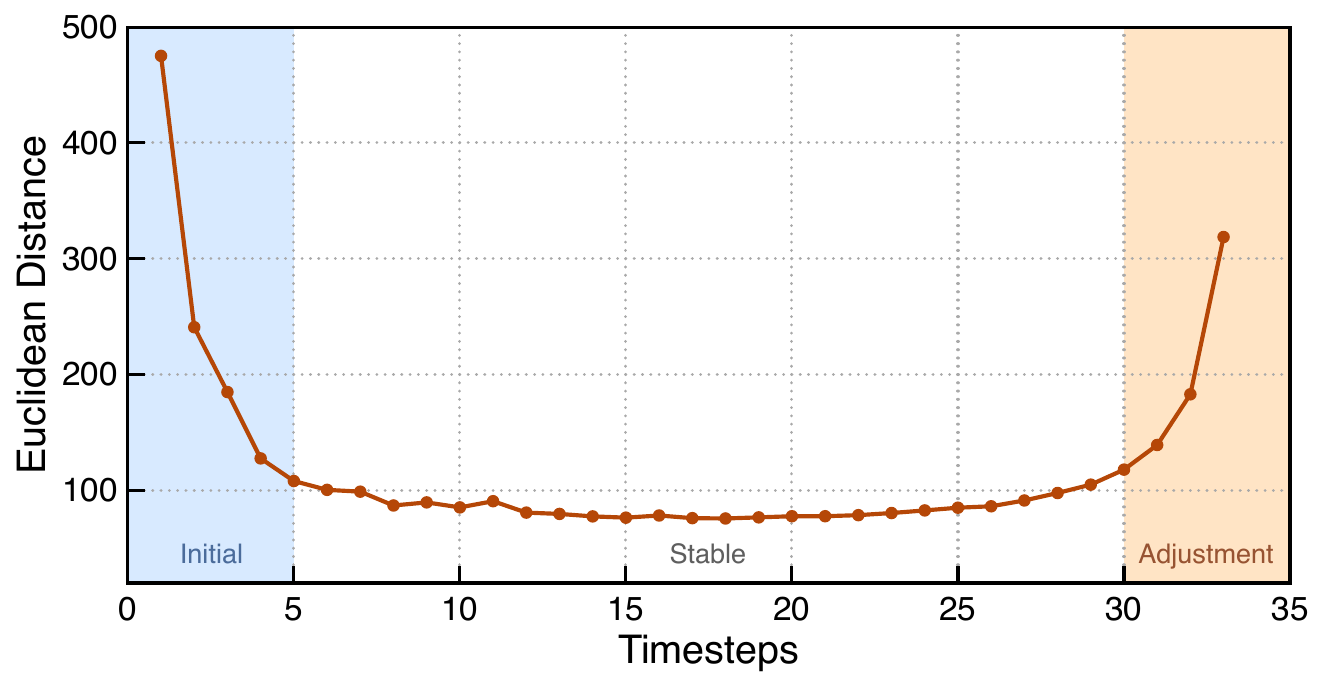}
\vspace{-6mm}
\caption{Euclidean distance between adjacent velocity predictions. A stable phase with minimal changes enables caching and reuse.}
\label{fig:velocity_smoothness}
\end{figure}

As shown in Figure~\ref{fig:velocity_smoothness}, measuring the Euclidean distance $\|v_{t+1} - v_t\|$ across timesteps reveals three distinct phases: (1) an \textit{initial phase} with rapidly changing predictions, (2) a \textit{stable phase} with minimal variation, and (3) a \textit{final adjustment phase} for fine-grained refinement. This suggests that we can avoid re-computing $v_\theta$ at every step during the stable phase by reusing previously cached values.

\noindent \textbf{Three-Stage Caching Strategy.}
Based on this observation, we divide the $N$-step denoising into three stages: (1) \textbf{Initial Phase} ($t < t_s$): velocity predictions are computed at each step due to high variability; (2) \textbf{Stable Phase} ($t_s \leq t < t_f$): velocities are computed every $\alpha$ steps and cached values are reused in between; (3) \textbf{Adjustment Phase} ($t \geq t_f$): full computation resumes for final refinement. We detect the stable phase onset using a relative smoothness threshold:
\begin{equation}
\frac{\|v_t - v_{t+1}\|}{\|v_0 - v_1\|} < k.
\end{equation}
With $k=0.4$, $\alpha=8$, and $m=3$ adjustment steps, this strategy achieves an average \textbf{61.2\%} time reduction on RoboTwin 2.0 with negligible quality degradation, enabling efficient scaling of our augmentation pipeline.

\begin{figure}[t]
    \centering
    \includegraphics[width=1\linewidth]{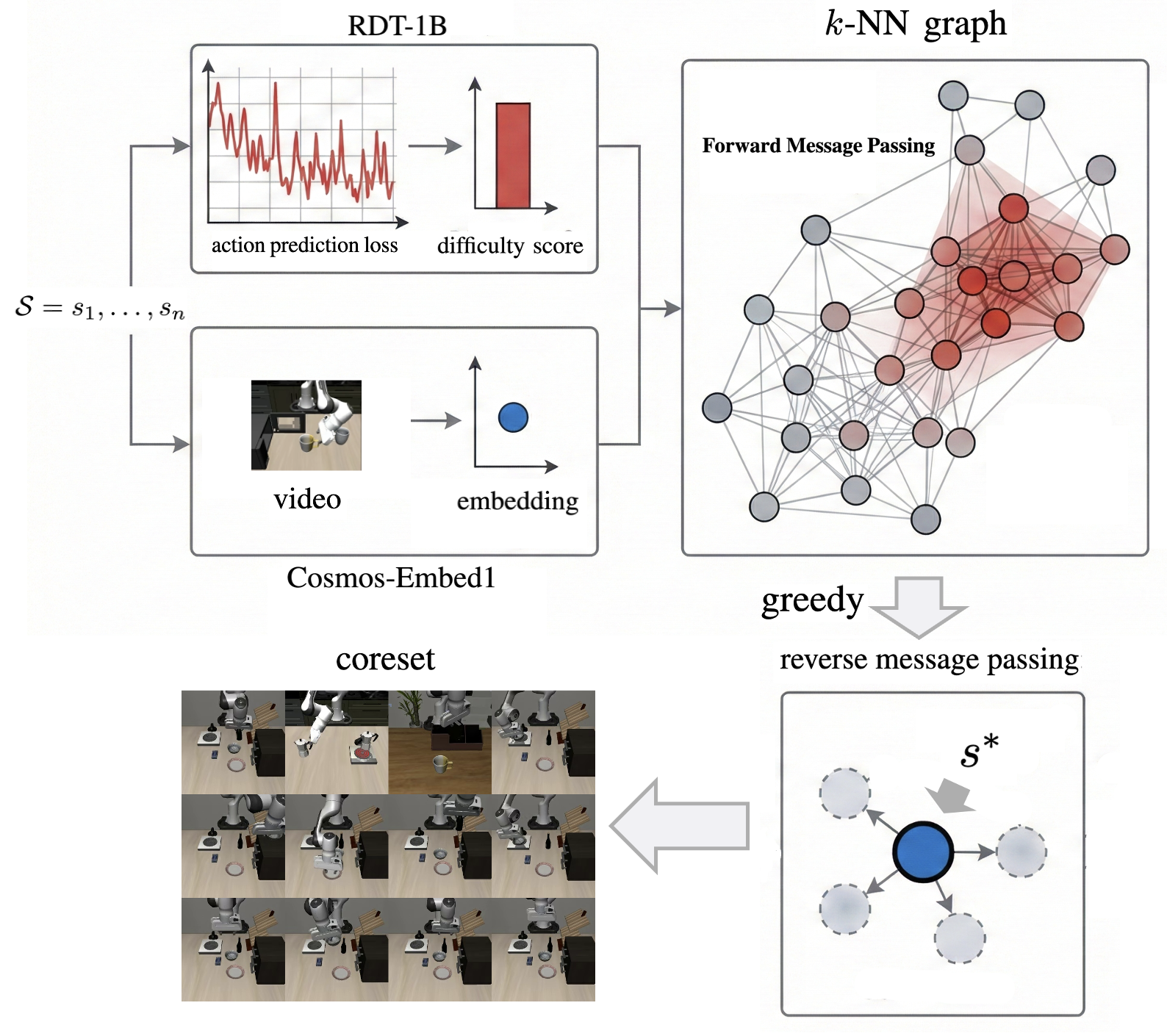}
    \vspace{-6mm}
    \caption{The proposed VLA coreset sampling algorithm.}
    \label{fig:corealg}
\end{figure}

\subsection{Selective Augmentation via Coreset Sampling}

Although our video augmentation framework enhances VLA training, applying it to all trajectories in a large-scale dataset is computationally prohibitive. To maximize augmentation utility under limited resources, we propose a coreset sampling strategy that identifies trajectories which are both challenging for policy models and visually diverse, as illustrated in Figure~\ref{fig:corealg}. Our method extends $\mathbb{D}^2$ Pruning~\cite{maharana2023d2} to embodied video data.

Given a set of simulated trajectories $\mathcal{S} = \{s_1, \dots, s_n\}$, our objective is to select a compact subset $\mathcal{S}' \subset \mathcal{S}$ that maximizes downstream benefit for sim-to-real transfer.

\paragraph{Difficulty Estimation via Policy Loss}
We quantify trajectory difficulty using a pre-trained VLA policy, such as RDT-1B~\cite{liu2024rdt}. For each trajectory $s_i$, we compute the average action prediction loss over a sampled subset of time steps $\mathcal{T}_i$ (to reduce computational cost):
\begin{equation}
x_i = \frac{1}{|\mathcal{T}_i|} \sum_{t \in \mathcal{T}i} \mathcal{L}_{\text{policy}}(s_i^{(t)}; \theta).
\end{equation}
A higher $x_i$ indicates that the policy finds the trajectory more challenging, which typically corresponds to edge cases or difficult configurations for manipulation.

\noindent \textbf{Diversity Estimation via Visual Embedding Topology}
To capture visual and semantic diversity, we extract trajectory embeddings and define neighbors using a state-of-the-art video representation model, Cosmos-Embed1~\cite{nvidia2025cosmosembed1}, denoted $\phi(s_i) \in \mathbb{R}^{768}$. We build a sparse $k$-NN graph over these embeddings with nodes $v_i = \phi(s_i)$ and edges weighted by an RBF kernel:
\begin{equation}
e_{i,j} = \exp(-\gamma_f \cdot \|v_i - v_j\|^2).
\end{equation}
This graph captures the topological structure of the visual space, enabling both forward and reverse message passing to score and prune redundant samples.

\noindent \textbf{Scoring via Forward Message Passing}
To prioritize clusters of difficult and non-redundant samples, we compute an updated difficulty score $x_i'$ that aggregates neighborhood difficulty:
\begin{equation}
x_i' = x_i + \sum_{j \in \mathcal{N}(i)} e_{i,j} \cdot x_j.
\end{equation}
This promotes samples that are not only hard in isolation but are also surrounded by similarly hard cases, emphasizing policy failure regions in the data.

\noindent \textbf{Coreset Construction with Redundancy Suppression}
We perform greedy selection using $x_i'$ scores and penalize neighbors of selected samples to avoid redundancy:
\begin{enumerate}
\item Select the trajectory with the highest score $s^* = \arg\max_{i\in\mathcal{S}_\text{remaining}}x_i'$, and add $s^*$ to selected coreset $\mathcal{S}'$.
\item Suppress scores of visually similar neighbors using reverse message passing:  
\[
x_j' \leftarrow x_j' - \exp(-\gamma_r \cdot \|v_{s^*} - v_j\|^2) \cdot x_{s^*}',
\]
where $\gamma_r$ controls the suppression radius.
\end{enumerate}
This iterative procedure balances sample hardness and coverage, yielding trajectories that are visually diverse and strategically valuable for sim-to-real transfer. By targeting augmentation toward this curated coreset, we achieve computational savings while preserving augmentation benefits. The details of hyperparameters above are as follows: $k=5, \gamma_r=0.1, \gamma_f=0.1$.

\section{Experiments}

\begin{figure}[htbp]
    \centering
    \includegraphics[width=1\linewidth]{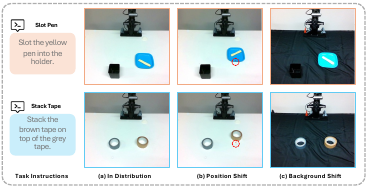}
    \caption{Two manipulation tasks (Slot Pen and Stack Tape) under three test conditions: (a) In-Distribution, (b) Position Shift, and (c) Background Shift.}
    \label{fig:distribution_shifts}
    \vspace{2mm}
\end{figure}

In the experiments below, first we prove the effectiveness of our augmentation method on single-task from Robotwin 2.0. Then, we move on to large scale learning using coreset sampling policy. To verify the advantages of our method  under environment with specific disturbances, we evaluate on LIBERO-Plus. Finally, to study the validity of our method in real word experiment, we design tasks and evaluate with an AgileX Piper robot arm under distributed environments, as shown in Figure \ref{fig:distribution_shifts}.  

\begin{table}[h]
\centering
\caption{Performance (\%) comparison of Hard and Easy scenarios of Robotwin 2.0~\cite{chen2025robotwin} - single-task learning. We train RDT~\cite{liu2024rdt} with the original simulation data (Ori.) \textit{vs.} augmented data (Aug.).}
\label{tab:rdt-comparison}
\setlength{\tabcolsep}{1.3mm}
\begin{tabular}{l|ll|ll}
\toprule
\multirow{2}{*}{\textbf{Task Name}} & \multicolumn{2}{c|}{\textbf{Hard}} & \multicolumn{2}{c}{\textbf{Easy}}\\
~ & Ori. & Aug. & Ori. & Aug.\\
\midrule
adjust\_bottle      & 72.0 & 82.0 \inc{10.0} & 74.0 & 90.0 \inc{16.0} \\
beat\_block\_hammer & 36.0 & 48.0 \inc{12.0} & 76.0 & 84.0 \inc{12.0} \\
pick\_dual\_bottles & 14.0 & 20.0 \inc{6.0} & 40.0 & 36.0 \dec{4.0}\\
place\_burger\_fries & 26.0 & 38.0 \inc{12.0} & 46.0 & 54.0 \inc{8.0}\\
open\_laptop        & 30.0 & 44.0 \inc{14.0} & 56.0 & 56.0\\
move\_can\_pot      & 12.0 & 18.0 \inc{6.0}& 26.0 & 34.0 \inc{8.0}\\
rotate\_qrcode      & 4.0 & 8.0 \inc{4.0} & 4.0 & 10.0 \inc{6.0}\\
grab\_roller        & 40.0 & 50.0 \inc{10.0} & 72.0 & 74.0 \inc{2.0}\\
\midrule
average             & 29.0 & 39.0 \inc{10.0} & 49.0 & 55.0 \inc{6.0} \\
\bottomrule
\end{tabular}
\end{table}

\subsection{Experimental Results on Robotwin 2.0}\label{exp:robotwin}
Following the experimental framework established by Robotwin 2.0~\cite{chen2025robotwin}, this study utilizes the Aloha AgileX dual-arm robot system. For the strategy model selection, we adopted both RDT-1B~\cite{liu2024rdt}  ACT~\cite{zhao2023learning} and $\pi_0$~\cite{black2410pi0} architectures, both fine-tuned based on officially pre-trained weights.\par

\textbf{\uline{(1) Experiments on single-task learning.}} We use 50 expert demonstration trajectories in a clean environment to train 10 tasks independently as the original baseline for single-task learning. While for our method, we extend augmented first-person perspective video data of all the trajectories, then train the same model for the same steps. We conducted 50 tests on each task under both ``Easy'' and ``Hard'' settings. The ``Easy'' mode corresponds to a \textit{clean} environment, while the ``Hard'' mode introduces \textit{domain-randomized} factors, including clutter, lighting, textures, and height variations.

\textbf{Results.} Table~\ref{tab:rdt-comparison} reports the performance of RDT-1B. Compared to the original baseline, our method yields substantial gains, most notably a 10\% improvement in ``Hard'', which includes domain-randomized factors and better reflects real-world complexity. Even in the clean easy scenario, we observe a 6\% increase, highlighting our method's ability to enhance generalization and mitigate overfitting.

Additional results in the Appendix further validate our approach. We report an average increase of 2.0 in $\pi_0$ (Table~\ref{tab:pi0-comparison}) and 3.0 in ACT (Table~\ref{tab:act-comparison}) under the hard setting of Robotwin 2.0 single-task learning, confirming the robustness of our augmentation strategy across different VLA policies.

\textbf{\uline{(2) Experiments on multi-task learning}}. We expanded the experimental scope to 32 diverse robot operation tasks with 300 trajectories on each task, resulting in a larger amount of $9600$ training trajectories. For our method, we adopt coreset sampling algorithm to select 10\% key samples in the full set, resulting in $9600$ training samples with 90\% original and 10\% augmented. When training, both experiments utilize the official pre-trained weights of RDT-1B and undergo large-scale multi-task joint training (100k steps). For evaluation metrics, we evaluated nine representative tasks for $50$ times in ``Hard'' scenario.

\begin{table}[htbp]
\centering
\caption{Performance (\%) comparison of Hard scenarios of Robotwin 2.0~\cite{chen2025robotwin} - multi-task learning. We train RDT~\cite{liu2024rdt} with the original simulation data (Ori.) \textit{vs.} augmented data (Aug.).}
\label{tab:aug-comparison}
\setlength{\tabcolsep}{2mm}
\begin{tabular}{l|cc}
\toprule
\textbf{Task Name} & \textbf{Ori.} & \textbf{Aug.} \\
\midrule
adjust\_bottle            & 20.0 & 56.0\inc{36.0} \\
beat\_block\_hammer       & 22.0 & 30.0\inc{8.0} \\
place\_burger\_fries      & 22.0 & 18.0\dec{4.0} \\
pick\_dual\_bottles       & 6.0 & 10.0 \inc{4.0}\\
place\_object\_basket     & 6.0 & 14.0 \inc{8.0} \\
move\_can\_pot           & 8.0 & 16.0 \inc{8.0} \\
stack\_bowls\_three      & 20.0 & 22.0 \inc{2.0} \\
click\_alarmclock        & 48.0 & 52.0 \inc{4.0} \\
move\_playingcard\_away  & 16.0 & 22.0 \inc{6.0} \\
open\_laptop             & 60.0 & 66.0 \inc{6.0} \\
\midrule
average                 & 23.0      &  31.0 \inc{8.0}    \\
\bottomrule
\end{tabular}
\vspace{-2mm}
\end{table}

\textbf{Results}. As shown in Table~\ref{tab:aug-comparison}, despite having 300 training trajectories per task, the baseline model struggles to generalize in the highly repetitive and visually uniform Robotwin environment. In contrast, augmenting just 10\% of the data with our method yields a significant performance boost. This demonstrates that our approach provides valuable visual and semantic diversity that the original dataset lacks, effectively regularizing the model and enhancing generalization, even under strong simulation constraints.

\subsection{Experiments on LIEBRO-Plus and LIBERO}\label{exp:libero}
To validate the effectiveness and generality of our proposed method on widely accepted public benchmarks, we selected LIBERO~\cite{liu2023libero}, and LIBERO-Plus~\cite{fei2025libero} to conduct experiments using $\pi_0$~\cite{black2410pi0} and $\pi_{0.5}$~\cite{intelligence2504pi0}.

We train all the models using the default standard configuration of OpenPi~\footnote{https://github.com/Physical-Intelligence/openpi}. For the baseline, we finetune the pretrained models with original dataset. While for our method, we augment a coreset of $50\%$ samples under two data compositions: (1) mixture strategy for mixing the augmented samples and all the original samples together; (2) replacement strategy for replacing the selected coreset with augmented ones.

\begin{table}[htbp]
\centering
\caption{
Performance (\%) comparison of spatial suite of LIBERO-Plus~\cite{fei2025libero} using $\pi_0$ and $\pi_{0.5}$.
}
\label{tab:openpi-comparison}
\setlength{\tabcolsep}{1.3mm}
\begin{tabular}{l|ll|ll}
\toprule
\multirow{2}{*}{\textbf{Perturbation Types}} & \multicolumn{2}{c|}{\textbf{$\pi_0$}} & \multicolumn{2}{c}{\textbf{$\pi_{0.5}$}}\\
~ & Ori. & Aug. & Ori. & Aug.\\
\midrule
light conditions     & 75.0 & 78.7 \inc{3.7} & 94.5 & 97.9 \inc{3.2} \\
objects layout & 69.6 & 86.2\inc{16.6} & 97.9 & 97.4 \dec{0.5} \\
background textures & 81.1 & 87.6 \inc{6.5} & 95.7 & 95.3 \dec{0.4}\\
sensor noise & 19.9 & 18.2\dec{1.7}& 91.2 & 93.4 \inc{2.2}\\
language instructions  &37.9 & 55.9\inc{22.0}  & 90.0 & 90.0 \\
robot initial states     & 10.3 & 6.3\dec{4.0} & 82.6 & 84.9 \inc{2.3}\\
camera view points     & 21.3 & 15.2\dec{6.1}  & 79.3 & 79.5 \inc{0.2} \\
\midrule
average             & 42.7 & 47.8 \inc{5.1} & 89.8 & 90.8 \inc{1.0} \\
\bottomrule
\end{tabular}
\vspace{2mm}
\end{table}

\textbf{Results}. After 30k-step training, We evaluate on both $\pi_{0}$ (mixture strategy) and $\pi_{0.5}$ (replacement strategy) on the spatial suite of LIEBRO-plus, which contains \textbf{2402}  evaluation settings. According to the results in Table~\ref{tab:openpi-comparison}, especially on $\pi_0$ 's performance comparison, is shows that adopting our augmentation method along with training strategy can enhance models' ability in simulated environments with disturbances like light conditions' changing. The slightly drop on camera viewpoints and robot initial states stem from a limitation of the current method: it mainly augments appearance, while these perturbations are geometry- and viewpoint-dependent. The results on $pi_{0.5}$ also denote that even with high performance model, using our method can still lead to obvious increase in some disturbance dimensions like sensor noise, light conditions and robot initial states.

Additionally, we also evaluate on LIBERO and observe a slight performance drop when training with augmented data: $\pi_0$ decreases by an average of 0.2 and $\pi_{0.5}$ by 0.5 across four task suites. This is expected, as LIBERO’s evaluation settings are nearly identical to its original training distribution—making aggressive augmentation introduce distributional disturbances rather than helpful diversity. Full results are provided in Table~\ref{tab:libero-comparison} (Appendix).

We further compare mixture \textit{vs.} replacement strategies for incorporating augmented data. Interestingly, $\pi_0$ performs best under the mixture strategy, while $\pi_{0.5}$ sees greater gains with replacement. This suggests that stronger pretrained models like $\pi_{0.5}$ benefit more from challenging augmentations that shift the training distribution, while weaker models prefer a balance between clean and augmented samples. See Table~\ref{tab:openpi-comparison_strategy} in the Appendix for details.

\subsection{Real-World Experiments and Results}

To validate the effectiveness of our video augmentation framework in bridging the sim-to-real gap, we conduct comprehensive experiments on a physical robotic platform with diverse out-of-distribution (OOD) test scenarios.

\noindent \textbf{Experimental setup.}
We deploy our method on an AgileX Piper manipulator using two state-of-the-art VLA model variants: $\pi_0$~\cite{black2410pi0} and $\pi_{0.5}$~\cite{intelligence2504pi0}. We evaluate on two representative manipulation tasks: (1) \textit{Stack Tape}: ``stacking a brown tape roll on top of another brown tape''; and (2) \textit{Slot Pen}: ``inserting a yellow pen into a holder''. To systematically evaluate generalization capability, we design three test conditions with increasing levels of distribution shift, namely In-Distribution, Position Shift (OOD), and Background Shift (OOD). For details, see Figure~\ref{fig:distribution_shifts} and Section \ref{appdix:real_world} in Appendix.

\noindent \textbf{Results.}
Table~\ref{tab:real_world_aug} presents the real-world experimental results across all tasks and test conditions. Our video augmentation method demonstrates consistent and substantial improvements over the baselines across both models.

\begin{table}[htbp]
\centering
\caption{Real-world results showing success counts (out of 10 trials) under three test conditions.}
\label{tab:real_world_aug}
\setlength{\tabcolsep}{1.0mm}
\begin{tabular}{l|cc|cc|cc|c}
\toprule
\multirow{2}{*}{\textbf{Method}} & \multicolumn{2}{c|}{\textbf{In-Dist.}} & \multicolumn{2}{c|}{\textbf{Position}} & \multicolumn{2}{c|}{\textbf{Background}} & \multirow{2}{*}{\textbf{Avg.}}\\
~ & Slot. & Stack. & Slot. & Stack. & Slot. & Stack. & \\
\midrule
$\pi_{0.5}$ & 9/10 & 10/10 & 5/10 & 4/10 & 4/10 & 4/10 & 60\% \\
+ Ours & \textbf{10/10} & \textbf{10/10} & \textbf{7/10} & \textbf{6/10} & \textbf{6/10} & \textbf{5/10} & \textbf{73\%} \\
\midrule
$\pi_{0}$ & 9/10 & 9/10 & 3/10 & 5/10 & 6/10 & 4/10 & 60\% \\
+ Ours & \textbf{10/10} & \textbf{10/10} & \textbf{5/10} & \textbf{8/10} & \textbf{5/10} & \textbf{7/10} & \textbf{75\%} \\
\bottomrule
\end{tabular}
\vspace{4mm}
\end{table}

\noindent \textbf{Analysis.}
Several key observations emerge:

\noindent \textit{(1) Consistent improvements across VLA architectures.} Our model-agnostic method yields 13\% and 15\% absolute gains in average success rates for $\pi_{0.5}$ and $\pi_0$, respectively. This validates that the visual diversity from our video transfer pipeline benefits various VLA backbones without requiring architecture-specific modifications.

\noindent \textit{(2) Enhanced robustness to spatial perturbations.} Significant gains under position shifts (e.g., $\pi_0$ on Stack Tape: 5/10 $\rightarrow$ 8/10) suggest the model learns generalizable spatial reasoning rather than memorizing fixed object configurations or action sequences.

\noindent \textit{(3) Improved visual domain generalization.} Improved performance under background shifts (e.g., $\pi_0$ on Stack Tape: 4/10 $\rightarrow$ 7/10) indicates that our synthetic diversity fosters robust feature representations that are resilient to task-irrelevant visual changes.

\noindent \textit{(4) Maintained in-distribution performance.} Augmented models maintain near-perfect in-distribution success rates (9--10/10), confirming that our pipeline preserves task-relevant semantics and action fidelity while introducing beneficial diversity.

\subsection{Ablation Study}

\textbf{Video augmentation with or without velocity cache.} To find the influence of using cache based acceleration on augmented videos, we train RDT-1B on four tasks from Robotwin 2.0 - single-task learning with or without velocity cache. As the results summarized in Table~\ref{tab:rdt-acc_comparsion}, both cases obtain similar accuracies and achieve significant improvements over the original baseline. We also compare their visual quality in Figure~\ref{fig:compareacc1} and Figure~\ref{fig:compareacc2} of Appendix. Although the cache acceleration technique introduces minor degradation in texture details at the pixel level, 
our method successfully preserves environmental semantic information essential for downstream policy learning, significantly reducing generation overhead without compromising the effectiveness of the augmented data.

\begin{table}[htbp]
\centering
\caption{Comparison of RDT-1B model finetuned using augmented data with or without velocity cache acceleration on Robotwin 2.0.
}
\label{tab:rdt-acc_comparsion}
\setlength{\tabcolsep}{1.3mm}
\begin{tabular}{l|ll|ll}
\toprule
\multirow{2}{*}{\textbf{Task Name}} & \multicolumn{2}{c|}{\textbf{Hard}} & \multicolumn{2}{c}{\textbf{Easy}}\\
~ & w/ acc. & w/o acc.& w/ acc. & w/o acc.\\
\midrule
move\_can\_pot      & 16.0 & 18.0  &34.0 & 34.0  \\
open\_laptop        & 42.0 & 44.0  &54.0 & 56.0  \\
rotate\_qrcode       & 10.0 & 8.0  &8.0 & 10.0 \\
place\_burger\_fries & 38.0 & 38.0 &54.0 & 52.0 \\
\midrule
average             & 26.5 & 27.0 & 37.5& 38.0 \\
\bottomrule
\end{tabular}
\vspace{4mm}
\end{table}\par

\begin{figure}[h!]
    \centering
    \includegraphics[width=1\linewidth]{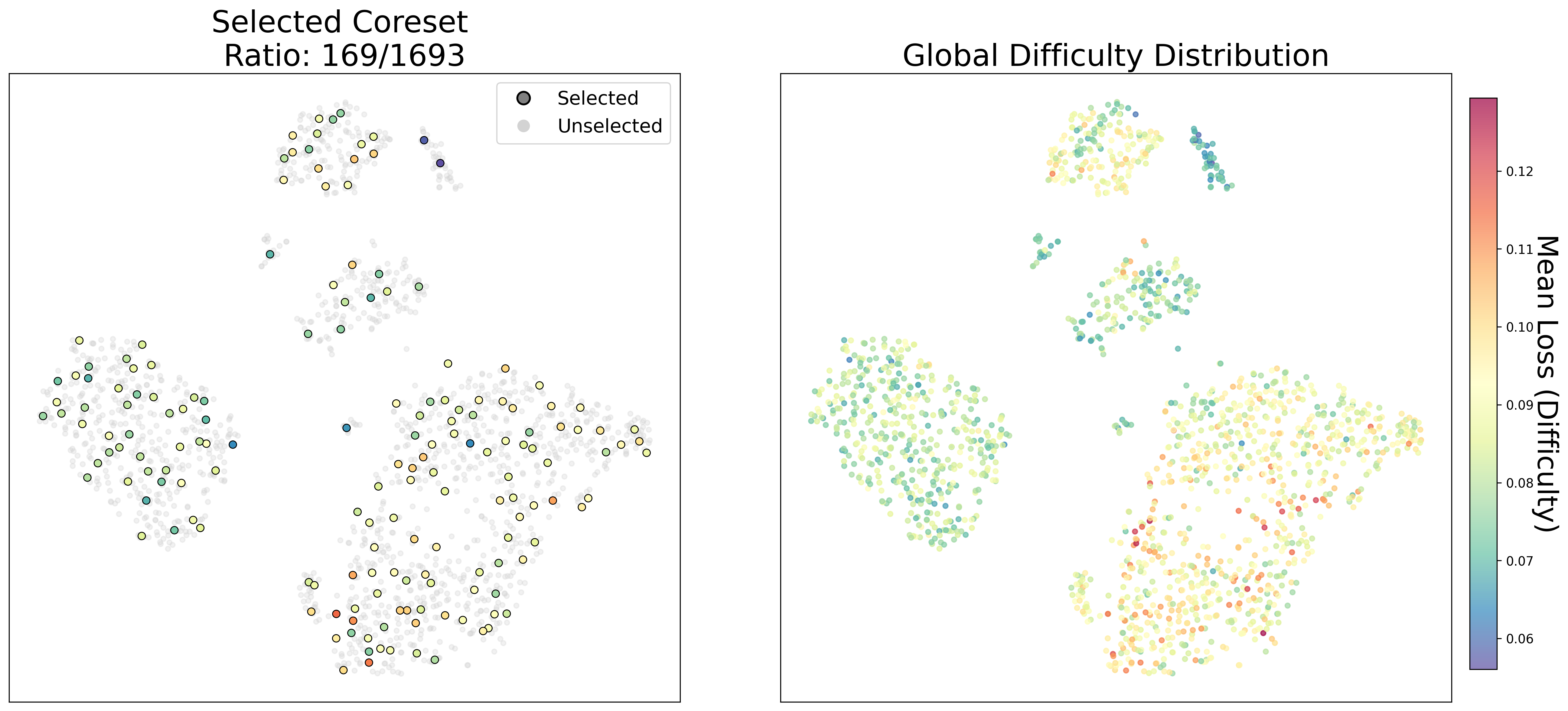}
    \caption{Visualization of coreset sampling on the LIBERO training dataset with a 10\% sampling budget. Right: the global difficulty distribution, where the color spectrum represents the mean policy loss (redder colors indicate higher difficulty). Left: the selected coreset overlaid on the full dataset. }
    \label{fig:Visualization of coreset selection}
    \vspace{4mm}
\end{figure}

\textbf{Evaluation of coreset sampling.} Figure~\ref{fig:Visualization of coreset selection} visualizes the video embeddings using t-SNE. The original dataset exhibits distinct clusters, each corresponding to semantically different tasks or visual contexts, with varying difficulty levels. Our selected coreset samples are well distributed across these clusters, covering both common and rare regions of the data manifold. Compared to random sampling, our method avoids over-representing redundant trajectories and instead targets high-loss, underrepresented areas. This balanced selection ensures that the augmented data introduces both semantic richness and policy-critical challenge, reinforcing generalization while avoiding unnecessary redundancy.

To evaluate how the proportion of selected samples affects performance, we conduct an ablation
using $\pi_0$ across 796 evaluation settings from LIBERO-Plus’s spatial suite. 

\begin{figure}[h!]
    \centering
    \includegraphics[width=\linewidth]{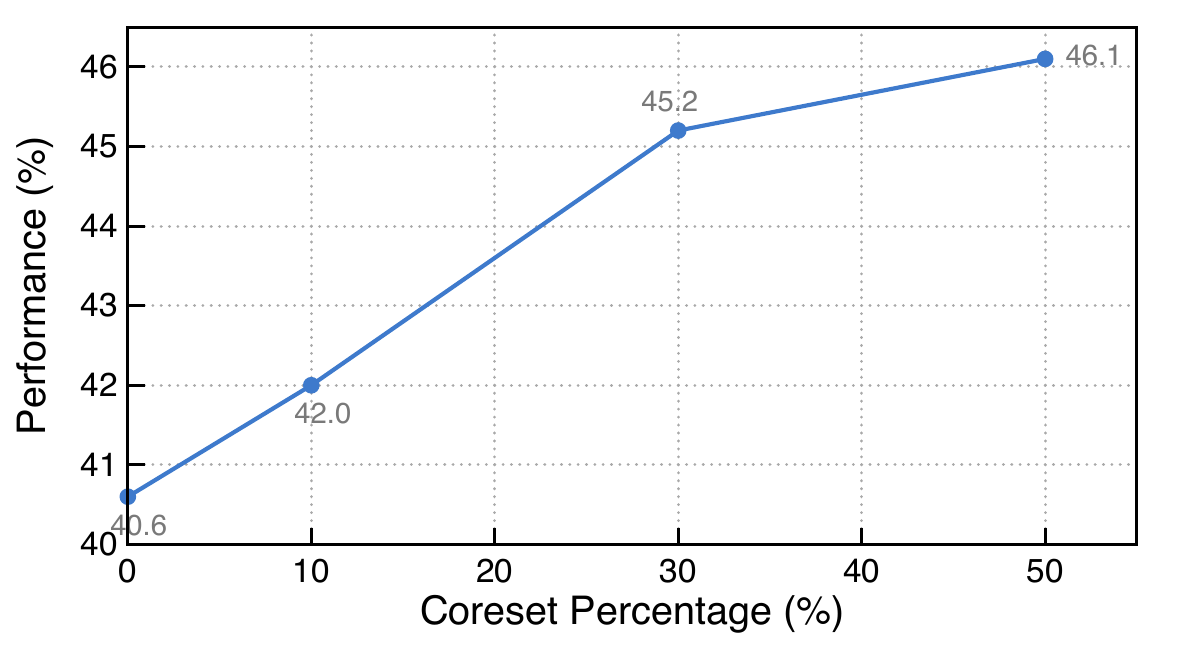}
    \caption{Performance comparison of different coreset sampling percentages on LIBERO-Plus spatial suite.}
    \label{fig:coreset_percentage}
\end{figure}

As shown in Figure~\ref{fig:coreset_percentage}, even a small augmentation ratio (10\%) yields a substantial improvement, with performance continuing to rise as more samples are included. However, the gain between 30\% and 50\% is only 0.9\%, indicating diminishing returns and highlighting the effectiveness of our coreset sampling in identifying critical and diverse samples.\par

\subsection{Video Transferring Performance Accessment}

To validate the efficacy of our proposed framework, we conducted a comparative study against the state-of-the-art baseline, RoboTransfer~\cite{liu2025robotransfer}. We evaluated performance across five representative tasks to assess both geometric consistency and semantic alignment. 
We employed two distinct sets of metrics to evaluate the synthesized video quality:\par
\textbf{Geometric Consistency}: We measured the depth fidelity between generated sequences and original sequences using Video Depth Anything~\cite{yang2024depth}. Metrics include the Root Mean Square Error (RMSE), Absolute Relative Error (Abs.Rel), and Squared Relative Error (Sq.Rel). Lower values indicate higher geometric consistency.\par
\textbf{Semantic Alignment}: We utilized VideoCLIP-XL~\cite{wang2024videoclip} to compute the prompt-video similarity, quantifying how well the generated video adheres to the provided textual instructions. Higher similarity scores indicate better semantic correspondence.\par
The quantitative results are summarized in Table~\ref{tab:robotransfer} and Table~\ref{tab:ours}. Our method consistently outperforms RoboTransfer across all evaluated tasks.

\begin{table}[h!]
\centering
\caption{RoboTransfer Results}
\label{tab:robotransfer}
\setlength{\tabcolsep}{1.3mm}

\begin{tabular}{lccccc}
\toprule
\textbf{task} & \textbf{RMSE} & \textbf{Abs.Rel} & \textbf{Sq.Rel} & \textbf{sim} & \textbf{time} \\
\midrule
adjust bottle & 0.46 & 0.37 & 0.39 & 21.5 & 340s \\
beat hammer & 0.30 & 0.28 & 0.16 & 22.3 & 368s \\
handover & 0.49 & 0.32 & 0.20 & 18.6 & 710s \\
hanging mug & 0.41 & 0.34 & 0.26 & 24.2 & 814s \\
pick bottles & 0.37 & 0.30 & 0.18 & 20.3 & 372s \\
\bottomrule
\end{tabular}
\end{table}

\begin{table}[h!]
\centering
\caption{Ours Results}
\label{tab:ours}
\setlength{\tabcolsep}{1.3mm}

\begin{tabular}{lccccc}
\toprule
\textbf{task} & \textbf{RMSE} & \textbf{Abs.Rel} & \textbf{Sq.Rel} & \textbf{sim} & \textbf{time} \\
\midrule
adjust bottle & 0.28 & 0.16 & 0.07 & 26.3 & 441s \\
beat hammer & 0.18 & 0.14 & 0.03 & 25.6 & 453s \\
handover block & 0.23 & 0.17 & 0.06 & 27.7 & 619s \\
hanging mug & 0.13 & 0.12 & 0.02 & 26.4 & 783s \\
pick bottles & 0.12 & 0.11 & 0.03 & 26.8 & 448s \\
\bottomrule
\end{tabular}
\end{table}

Regarding geometric consistency, we observed a substantial reduction in error rates. Notably, the RMSE improved from 0.46 to 0.28 in the "adjust bottle" task with Abs.Rel and Sq.Rel metrics demonstrating a 2–6× reduction across the board. Furthermore, our approach achieved superior semantic alignment, with prompt similarity scores consistently increasing from the 18.6–24.2 range to 25.6–27.7 while our method incurs a marginal increase in runtime—averaging.
\section{Conclusion}

We present a scalable video augmentation framework designed to address the generalization limitations of vision-language-action models trained on simulation. Our method introduces semantic and visual diversity into synthetic training data by combining structured caption rewriting, conditional video generation, efficient velocity-prediction caching, and coreset-based selective augmentation. Through  experiments on Robotwin, LIBERO, and LIBERO-Plus, we demonstrate consistent performance improvements and enhanced robustness to real-world perturbations. Our results highlight that when diverse and targeted, even modest augmentation is effective in improving model generalization.

\section*{Acknowledgements}
This work was supported by the National Natural Science Foundation of China under Grant 62506235.

\section*{Impact Statement}


\paragraph{Ethical Considerations}
This research focuses on enhancing the generalization of Vision-Language-Action models through efficient sim-to-real video augmentation. While aiming to reduce reliance on costly real-world data, we emphasize the need for ethical vigilance in deploying such technologies. Key considerations include ensuring transparency in generated data, mitigating biases inherited from simulations or language models, and maintaining alignment with human values in autonomous decision-making. We encourage ongoing community efforts to address these ethical challenges alongside technical progress.

\paragraph{Societal Implications}
By lowering barriers to robust robotic system development, this work could positively impact fields like industrial automation and assisted services. However, broader adoption may also introduce societal shifts, such as changes in labor dynamics or unintended use cases. We advocate for proactive dialogue to guide responsible innovation, ensuring benefits are widely distributed while risks are carefully managed. In line with standard practice, we defer detailed societal impact discussions to broader interdisciplinary efforts.

\nocite{langley00}

\bibliography{main}

@inproceedings{langley00,
 author    = {P. Langley},
 title     = {Crafting Papers on Machine Learning},
 year      = {2000},
 pages     = {1207--1216},
 editor    = {Pat Langley},
 booktitle     = {Proceedings of the 17th International Conference
              on Machine Learning (ICML 2000)},
 address   = {Stanford, CA},
 publisher = {Morgan Kaufmann}
}

@article{li2023videochat,
  title={VideoChat: Chat-Centric Video Understanding},
  author={Li, Kunchang and He, Yinan and Wang, Yi and Li, Yizhuo and Wang, Wenhai and Luo, Ping and Wang, Yali and Wang, Limin and Qiao, Yu},
  journal={CoRR},
  year={2023}
}

@article{xu2025egodemogen,
  title={Egodemogen: Novel egocentric demonstration generation enables viewpoint-robust manipulation},
  author={Xu, Yuan and Yang, Jiabing and Wang, Xiaofeng and Chen, Yixiang and Zhu, Zheng and Fang, Bowen and Huang, Guan and Chen, Xinze and Ye, Yun and Zhang, Qiang and others},
  journal={arXiv preprint arXiv:2509.22578},
  year={2025}
}

@article{dong2025emma,
  title={Emma: Generalizing real-world robot manipulation via generative visual transfer},
  author={Dong, Zhehao and Wang, Xiaofeng and Zhu, Zheng and Wang, Yirui and Wang, Yang and Zhou, Yukun and Wang, Boyuan and Ni, Chaojun and Ouyang, Runqi and Qin, Wenkang and others},
  journal={arXiv preprint arXiv:2509.22407},
  year={2025}
}

@article{yang2025qwen3,
  title={Qwen3 technical report},
  author={Yang, An and Li, Anfeng and Yang, Baosong and Zhang, Beichen and Hui, Binyuan and Zheng, Bo and Yu, Bowen and Gao, Chang and Huang, Chengen and Lv, Chenxu and others},
  journal={arXiv preprint arXiv:2505.09388},
  year={2025}
}

@article{ali2025world,
  title={World simulation with video foundation models for physical ai},
  author={Ali, Arslan and Bai, Junjie and Bala, Maciej and Balaji, Yogesh and Blakeman, Aaron and Cai, Tiffany and Cao, Jiaxin and Cao, Tianshi and Cha, Elizabeth and Chao, Yu-Wei and others},
  journal={arXiv preprint arXiv:2511.00062},
  year={2025}
}

@article{alhaija2025cosmos,
  title={Cosmos-transfer1: Conditional world generation with adaptive multimodal control},
  author={Alhaija, Hassan Abu and Alvarez, Jose and Bala, Maciej and Cai, Tiffany and Cao, Tianshi and Cha, Liz and Chen, Joshua and Chen, Mike and Ferroni, Francesco and Fidler, Sanja and others},
  journal={arXiv preprint arXiv:2503.14492},
  year={2025}
}

@article{team2025gigabrain,
  title={Gigabrain-0: A world model-powered vision-language-action model},
  author={Team, GigaBrain and Ye, Angen and Wang, Boyuan and Ni, Chaojun and Huang, Guan and Zhao, Guosheng and Li, Haoyun and Li, Jie and Zhu, Jiagang and Feng, Lv and others},
  journal={arXiv preprint arXiv:2510.19430},
  year={2025}
}

@article{team2025gigaworld,
  title={Gigaworld-0: World models as data engine to empower embodied ai},
  author={Team, GigaWorld and Ye, Angen and Wang, Boyuan and Ni, Chaojun and Huang, Guan and Zhao, Guosheng and Li, Haoyun and Zhu, Jiagang and Li, Kerui and Xu, Mengyuan and others},
  journal={arXiv preprint arXiv:2511.19861},
  year={2025}
}

@article{yang2024depth,
  title={Depth anything v2},
  author={Yang, Lihe and Kang, Bingyi and Huang, Zilong and Zhao, Zhen and Xu, Xiaogang and Feng, Jiashi and Zhao, Hengshuang},
  journal={Advances in Neural Information Processing Systems},
  volume={37},
  pages={21875--21911},
  year={2024}
}

@article{ravi2024sam,
  title={Sam 2: Segment anything in images and videos},
  author={Ravi, Nikhila and Gabeur, Valentin and Hu, Yuan-Ting and Hu, Ronghang and Ryali, Chaitanya and Ma, Tengyu and Khedr, Haitham and R{\"a}dle, Roman and Rolland, Chloe and Gustafson, Laura and others},
  journal={arXiv preprint arXiv:2408.00714},
  year={2024}
}

@article{wang2024videoclip,
  title={Videoclip-xl: Advancing long description understanding for video clip models},
  author={Wang, Jiapeng and Wang, Chengyu and Huang, Kunzhe and Huang, Jun and Jin, Lianwen},
  journal={arXiv preprint arXiv:2410.00741},
  year={2024}
}

@article{wan2025wan,
  title={Wan: Open and advanced large-scale video generative models},
  author={Wan, Team and Wang, Ang and Ai, Baole and Wen, Bin and Mao, Chaojie and Xie, Chen-Wei and Chen, Di and Yu, Feiwu and Zhao, Haiming and Yang, Jianxiao and others},
  journal={arXiv preprint arXiv:2503.20314},
  year={2025}
}

@article{jiang2025vace,
  title={Vace: All-in-one video creation and editing},
  author={Jiang, Zeyinzi and Han, Zhen and Mao, Chaojie and Zhang, Jingfeng and Pan, Yulin and Liu, Yu},
  journal={arXiv preprint arXiv:2503.07598},
  year={2025}
}

@inproceedings{liu2025timestep,
  title={Timestep Embedding Tells: It's Time to Cache for Video Diffusion Model},
  author={Liu, Feng and Zhang, Shiwei and Wang, Xiaofeng and Wei, Yujie and Qiu, Haonan and Zhao, Yuzhong and Zhang, Yingya and Ye, Qixiang and Wan, Fang},
  booktitle={Proceedings of the Computer Vision and Pattern Recognition Conference},
  pages={7353--7363},
  year={2025}
}

@article{zhou2025less,
  title={Less is Enough: Training-Free Video Diffusion Acceleration via Runtime-Adaptive Caching},
  author={Zhou, Xin and Liang, Dingkang and Chen, Kaijin and Feng, Tianrui and Chen, Xiwu and Lin, Hongkai and Ding, Yikang and Tan, Feiyang and Zhao, Hengshuang and Bai, Xiang},
  journal={arXiv preprint arXiv:2507.02860},
  year={2025}
}

@article{lipman2022flow,
  title={Flow matching for generative modeling},
  author={Lipman, Yaron and Chen, Ricky TQ and Ben-Hamu, Heli and Nickel, Maximilian and Le, Matt},
  journal={arXiv preprint arXiv:2210.02747},
  year={2022}
}

@article{chen2025robotwin,
  title={Robotwin 2.0: A scalable data generator and benchmark with strong domain randomization for robust bimanual robotic manipulation},
  author={Chen, Tianxing and Chen, Zanxin and Chen, Baijun and Cai, Zijian and Liu, Yibin and Li, Zixuan and Liang, Qiwei and Lin, Xianliang and Ge, Yiheng and Gu, Zhenyu and others},
  journal={arXiv preprint arXiv:2506.18088},
  year={2025}
}

@article{maharana2023d2,
  title={D2 pruning: Message passing for balancing diversity and difficulty in data pruning},
  author={Maharana, Adyasha and Yadav, Prateek and Bansal, Mohit},
  journal={arXiv preprint arXiv:2310.07931},
  year={2023}
}

@article{liu2024rdt,
  title={Rdt-1b: a diffusion foundation model for bimanual manipulation},
  author={Liu, Songming and Wu, Lingxuan and Li, Bangguo and Tan, Hengkai and Chen, Huayu and Wang, Zhengyi and Xu, Ke and Su, Hang and Zhu, Jun},
  journal={arXiv preprint arXiv:2410.07864},
  year={2024}
}

@misc{nvidia2025cosmosembed1,
    title = {Cosmos-Embed1: A Joint Video-Text Embedder for Physical AI},
    author = {NVIDIA and Ferroni, Francesco and Chattopadhyay, Prithvijit and Heinrich, Greg and Ranzinger, Mike and Amoroso, Roberto and Luo, Alice and Wang, Andrew and Liu, Ming-Yu},
    year = {2025},
    url = {https://research.nvidia.com/labs/dir/cosmos-embed1},
    }

@inproceedings{rombach2022high,
  title={High-resolution image synthesis with latent diffusion models},
  author={Rombach, Robin and Blattmann, Andreas and Lorenz, Dominik and Esser, Patrick and Ommer, Bj{\"o}rn},
  booktitle={Proceedings of the IEEE/CVF conference on computer vision and pattern recognition},
  pages={10684--10695},
  year={2022}
}

@article{blattmann2023stable,
  title={Stable video diffusion: Scaling latent video diffusion models to large datasets},
  author={Blattmann, Andreas and Dockhorn, Tim and Kulal, Sumith and Mendelevitch, Daniel and Kilian, Maciej and Lorenz, Dominik and Levi, Yam and English, Zion and Voleti, Vikram and Letts, Adam and others},
  journal={arXiv preprint arXiv:2311.15127},
  year={2023}
}

@article{brooks2024video,
  title={Video generation models as world simulators},
  author={Brooks, Tim and Peebles, Bill and Holmes, Connor and DePue, Will and Guo, Yufei and Jing, Li and Schnurr, David and Taylor, Joe and Luhman, Troy and Luhman, Eric and others},
  journal={OpenAI Blog},
  volume={1},
  number={8},
  pages={1},
  year={2024}
}

@article{fang2025rebot,
  title={Rebot: Scaling robot learning with real-to-sim-to-real robotic video synthesis},
  author={Fang, Yu and Yang, Yue and Zhu, Xinghao and Zheng, Kaiyuan and Bertasius, Gedas and Szafir, Daniel and Ding, Mingyu},
  journal={arXiv preprint arXiv:2503.14526},
  year={2025}
}

@article{wang2025embodiedreamer,
  title={Embodiedreamer: Advancing real2sim2real transfer for policy training via embodied world modeling},
  author={Wang, Boyuan and Meng, Xinpan and Wang, Xiaofeng and Zhu, Zheng and Ye, Angen and Wang, Yang and Yang, Zhiqin and Ni, Chaojun and Huang, Guan and Wang, Xingang},
  journal={arXiv preprint arXiv:2507.05198},
  year={2025}
}

@inproceedings{wenzek2020ccnet,
  title={CCNet: Extracting high quality monolingual datasets from web crawl data},
  author={Wenzek, Guillaume and Lachaux, Marie-Anne and Conneau, Alexis and Chaudhary, Vishrav and Guzm{\'a}n, Francisco and Joulin, Armand and Grave, Edouard},
  booktitle={Proceedings of the twelfth language resources and evaluation conference},
  pages={4003--4012},
  year={2020}
}

@article{raffel2020exploring,
  title={Exploring the limits of transfer learning with a unified text-to-text transformer},
  author={Raffel, Colin and Shazeer, Noam and Roberts, Adam and Lee, Katherine and Narang, Sharan and Matena, Michael and Zhou, Yanqi and Li, Wei and Liu, Peter J},
  journal={Journal of machine learning research},
  volume={21},
  number={140},
  pages={1--67},
  year={2020}
}

@inproceedings{lee2022deduplicating,
  title={Deduplicating training data makes language models better},
  author={Lee, Katherine and Ippolito, Daphne and Nystrom, Andrew and Zhang, Chiyuan and Eck, Douglas and Callison-Burch, Chris and Carlini, Nicholas},
  booktitle={Proceedings of the 60th Annual Meeting of the Association for Computational Linguistics (Volume 1: Long Papers)},
  pages={8424--8445},
  year={2022}
}

@article{abbas2023semdedup,
  title={Semdedup: Data-efficient learning at web-scale through semantic deduplication},
  author={Abbas, Amro and Tirumala, Kushal and Simig, D{\'a}niel and Ganguli, Surya and Morcos, Ari S},
  journal={arXiv preprint arXiv:2303.09540},
  year={2023}
}

@article{gao2020pile,
  title={The pile: An 800gb dataset of diverse text for language modeling},
  author={Gao, Leo and Biderman, Stella and Black, Sid and Golding, Laurence and Hoppe, Travis and Foster, Charles and Phang, Jason and He, Horace and Thite, Anish and Nabeshima, Noa and others},
  journal={arXiv preprint arXiv:2101.00027},
  year={2020}
}

@article{penedo2023refinedweb,
  title={The RefinedWeb dataset for Falcon LLM: outperforming curated corpora with web data, and web data only},
  author={Penedo, Guilherme and Malartic, Quentin and Hesslow, Daniel and Cojocaru, Ruxandra and Cappelli, Alessandro and Alobeidli, Hamza and Pannier, Baptiste and Almazrouei, Ebtesam and Launay, Julien},
  journal={arXiv preprint arXiv:2306.01116},
  year={2023}
}

@article{xie2023data,
  title={Data selection for language models via importance resampling},
  author={Xie, Sang Michael and Santurkar, Shibani and Ma, Tengyu and Liang, Percy S},
  journal={Advances in Neural Information Processing Systems},
  volume={36},
  pages={34201--34227},
  year={2023}
}

@article{sorscher2022beyond,
  title={Beyond neural scaling laws: beating power law scaling via data pruning},
  author={Sorscher, Ben and Geirhos, Robert and Shekhar, Shashank and Ganguli, Surya and Morcos, Ari},
  journal={Advances in Neural Information Processing Systems},
  volume={35},
  pages={19523--19536},
  year={2022}
}

@article{black2410pi0,
  title={$\pi$0: A vision-language-action flow model for general robot control. CoRR, abs/2410.24164, 2024. doi: 10.48550},
  author={Black, Kevin and Brown, Noah and Driess, Danny and Esmail, Adnan and Equi, Michael and Finn, Chelsea and Fusai, Niccolo and Groom, Lachy and Hausman, Karol and Ichter, Brian and others},
  journal={arXiv preprint ARXIV.2410.24164}
}

@article{intelligence2504pi0,
  title={$\pi$0. 5: a vision-language-action model with open-world generalization, 2025},
  author={Intelligence, Physical and Black, Kevin and Brown, Noah and Darpinian, James and Dhabalia, Karan and Driess, Danny and Esmail, Adnan and Equi, Michael and Finn, Chelsea and Fusai, Niccolo and others},
  journal={URL https://arxiv. org/abs/2504.16054},
  volume={1},
  number={2},
  pages={3}
}

@article{fei2025libero,
  title={Libero-plus: In-depth robustness analysis of vision-language-action models},
  author={Fei, Senyu and Wang, Siyin and Shi, Junhao and Dai, Zihao and Cai, Jikun and Qian, Pengfang and Ji, Li and He, Xinzhe and Zhang, Shiduo and Fei, Zhaoye and others},
  journal={arXiv preprint arXiv:2510.13626},
  year={2025}
}

@inproceedings{li2023diffnas,
  title={Diffnas: Bootstrapping diffusion models by prompting for better architectures},
  author={Li, Wenhao and Su, Xiu and You, Shan and Wang, Fei and Qian, Chen and Xu, Chang},
  booktitle={2023 IEEE International Conference on Data Mining (ICDM)},
  pages={1121--1126},
  year={2023},
  organization={IEEE}
}

@article{liu2023libero,
  title={Libero: Benchmarking knowledge transfer for lifelong robot learning},
  author={Liu, Bo and Zhu, Yifeng and Gao, Chongkai and Feng, Yihao and Liu, Qiang and Zhu, Yuke and Stone, Peter},
  journal={Advances in Neural Information Processing Systems},
  volume={36},
  pages={44776--44791},
  year={2023}
}

@article{zhao2023learning,
  title={Learning fine-grained bimanual manipulation with low-cost hardware},
  author={Zhao, Tony Z and Kumar, Vikash and Levine, Sergey and Finn, Chelsea},
  journal={arXiv preprint arXiv:2304.13705},
  year={2023}
}

@misc{
li2025vlas,
title={{VLA}s Are Confined yet Capable of Generalizing to Novel Tasks},
author={Quanyi Li},
year={2025},
url={https://openreview.net/forum?id=jtrhwfgseW}
}

@article{zhou2025libero,
  title={LIBERO-PRO: Towards Robust and Fair Evaluation of Vision-Language-Action Models Beyond Memorization},
  author={Zhou, Xueyang and Xu, Yangming and Tie, Guiyao and Chen, Yongchao and Zhang, Guowen and Chu, Duanfeng and Zhou, Pan and Sun, Lichao},
  journal={arXiv preprint arXiv:2510.03827},
  year={2025}
}

@article{brohan2022rt,
  title={Rt-1: Robotics transformer for real-world control at scale},
  author={Brohan, Anthony and Brown, Noah and Carbajal, Justice and Chebotar, Yevgen and Dabis, Joseph and Finn, Chelsea and Gopalakrishnan, Keerthana and Hausman, Karol and Herzog, Alex and Hsu, Jasmine and others},
  journal={arXiv preprint arXiv:2212.06817},
  year={2022}
}

@article{ebert2021bridge,
  title={Bridge data: Boosting generalization of robotic skills with cross-domain datasets},
  author={Ebert, Frederik and Yang, Yanlai and Schmeckpeper, Karl and Bucher, Bernadette and Georgakis, Georgios and Daniilidis, Kostas and Finn, Chelsea and Levine, Sergey},
  journal={arXiv preprint arXiv:2109.13396},
  year={2021}
}

@inproceedings{karamcheti2024prismatic,
  title={Prismatic vlms: Investigating the design space of visually-conditioned language models},
  author={Karamcheti, Siddharth and Nair, Suraj and Balakrishna, Ashwin and Liang, Percy and Kollar, Thomas and Sadigh, Dorsa},
  booktitle={Forty-first International Conference on Machine Learning},
  year={2024}
}

@article{kim2024openvla,
  title={Openvla: An open-source vision-language-action model},
  author={Kim, Moo Jin and Pertsch, Karl and Karamcheti, Siddharth and Xiao, Ted and Balakrishna, Ashwin and Nair, Suraj and Rafailov, Rafael and Foster, Ethan and Lam, Grace and Sanketi, Pannag and others},
  journal={arXiv preprint arXiv:2406.09246},
  year={2024}
}

@misc{intelligence2025pi06vlalearnsexperience,
      title={$\pi^{*}_{0.6}$: a VLA That Learns From Experience}, 
      author={Physical Intelligence and Ali Amin and Raichelle Aniceto and Ashwin Balakrishna and Kevin Black and Ken Conley and Grace Connors and James Darpinian and Karan Dhabalia and Jared DiCarlo and Danny Driess and Michael Equi and Adnan Esmail and Yunhao Fang and Chelsea Finn and Catherine Glossop and Thomas Godden and Ivan Goryachev and Lachy Groom and Hunter Hancock and Karol Hausman and Gashon Hussein and Brian Ichter and Szymon Jakubczak and Rowan Jen and Tim Jones and Ben Katz and Liyiming Ke and Chandra Kuchi and Marinda Lamb and Devin LeBlanc and Sergey Levine and Adrian Li-Bell and Yao Lu and Vishnu Mano and Mohith Mothukuri and Suraj Nair and Karl Pertsch and Allen Z. Ren and Charvi Sharma and Lucy Xiaoyang Shi and Laura Smith and Jost Tobias Springenberg and Kyle Stachowicz and Will Stoeckle and Alex Swerdlow and James Tanner and Marcel Torne and Quan Vuong and Anna Walling and Haohuan Wang and Blake Williams and Sukwon Yoo and Lili Yu and Ury Zhilinsky and Zhiyuan Zhou},
      year={2025},
      eprint={2511.14759},
      archivePrefix={arXiv},
      primaryClass={cs.LG},
      url={https://arxiv.org/abs/2511.14759}, 
}

@article{fang2025intention,
  title={From intention to execution: Probing the generalization boundaries of vision-language-action models},
  author={Fang, Irving and Zhang, Juexiao and Tong, Shengbang and Feng, Chen},
  journal={arXiv preprint arXiv:2506.09930},
  year={2025}
}

@article{xu2025affordance,
  title={Affordance Field Intervention: Enabling VLAs to Escape Memory Traps in Robotic Manipulation},
  author={Xu, Siyu and Wang, Zijian and Wang, Yunke and Xia, Chenghao and Huang, Tao and Xu, Chang},
  journal={arXiv preprint arXiv:2512.07472},
  year={2025}
}

@article{pei2025action,
  title={Action-aware dynamic pruning for efficient vision-language-action manipulation},
  author={Pei, Xiaohuan and Chen, Yuxing and Xu, Siyu and Wang, Yunke and Shi, Yuheng and Xu, Chang},
  journal={arXiv preprint arXiv:2509.22093},
  year={2025}
}

@article{xu2025vla,
  title={VLA-Cache: Efficient Vision-Language-Action Manipulation via Adaptive Token Caching},
  author={Xu, Siyu and Wang, Yunke and Xia, Chenghao and Zhu, Dihao and Huang, Tao and Xu, Chang},
  journal={arXiv preprint arXiv:2502.02175},
  year={2025}
}

@article{liu2025robotransfer,
  title={Robotransfer: Geometry-consistent video diffusion for robotic visual policy transfer},
  author={Liu, Liu and Wang, Xiaofeng and Zhao, Guosheng and Li, Keyu and Qin, Wenkang and Zhu, Jiagang and Qiu, Jiaxiong and Zhu, Zheng and Huang, Guan and Su, Zhizhong},
  year={2025}
}
\bibliographystyle{icml2026}

\newpage
\appendix
\onecolumn

\section{Additional Experiments}

\subsection{Results of More Model Variants on RoboTwin 2.0}
For ACT and $\pi_0$, we carry out experiment on single task from RoboTwin 2.0 using 50 trajectories collected in clean environment for finetuning. The average increase on hard mode is 3.0 for ACT and 2.0 for $\pi_0$. These results show that our method is suitable for both VLA framework including pre-trained model (ACT) and finetuning model ($\pi_0$).
\begin{table}[htbp]
\centering
\caption{ACT: Performance (\%) comparison of the original simulation data (Ori.) \textit{vs.} augmented data (Aug.) on ACT~\cite{zhao2023learning} in Hard vs. Easy scenarios of Robotwin 2.0~\cite{chen2025robotwin}}
\label{tab:act-comparison}
\setlength{\tabcolsep}{1.3mm}
\begin{tabular}{l|ll|ll}
\toprule
\multirow{2}{*}{\textbf{Task Name}} & \multicolumn{2}{c|}{\textbf{Hard}} & \multicolumn{2}{c}{\textbf{Easy}}\\
~ & Ori. & Aug. & Ori. & Aug.\\
\midrule
adjust\_bottle      & 20.0 & 26.0 \inc{6.0} & 92.0 & 94.0 \inc{2.0} \\
beat\_block\_hammer & 2.0 & 2.0 & 56.0 & 52.0 \dec{4.0} \\
pick\_dual\_bottles & 0.0 & 4.0 \inc{4.0} & 30.0 & 34.0 \inc{4.0}\\
place\_burger\_fries & 0.0 & 0.0& 46.0 & 52.0 \inc{6.0}\\
open\_laptop        & 0.0 & 0.0  & 54.0 & 58.0 \inc{4.0}\\
move\_can\_pot      & 2.0 & 4.0 \inc{2.0}& 20.0 & 28.0 \inc{8.0}\\
rotate\_qrcode      & 0.0 & 0.0  & 2.0 & 2.0 \\
grab\_roller        & 22.0 & 32.0 \inc{10.0} & 94.0 & 90.0 \dec{4.0}\\
\midrule
average             & 6.0 & 9.0 \inc{3.0} & 49.0 & 51.0 \inc{2.0} \\
\bottomrule
\end{tabular}
\end{table}

\begin{table}[htbp]
\centering
\caption{$\pi_0$: Performance (\%) comparison of the original simulation data (Ori.) \textit{vs.} augmented data (Aug.) on $\pi0$~\cite{black2410pi0} in Hard vs. Easy scenarios of Robotwin 2.0~\cite{chen2025robotwin}}
\label{tab:pi0-comparison}
\setlength{\tabcolsep}{1.3mm}
\begin{tabular}{l|ll|ll}
\toprule
\multirow{2}{*}{\textbf{Task Name}} & \multicolumn{2}{c|}{\textbf{Hard}} & \multicolumn{2}{c}{\textbf{Easy}}\\
~ & Ori. & Aug. & Ori. & Aug.\\
\midrule
adjust\_bottle      & 74.0 & 78.0 \inc{4.0} & 90.0 & 92.0 \inc{2.0} \\
beat\_block\_hammer & 20.0 & 16.0\dec{4.0} & 42.0 & 48.0 \dec{6.0} \\
pick\_dual\_bottles & 12.0 & 10.0 \dec{2.0} & 56.0 & 60.0 \inc{4.0}\\
place\_burger\_fries & 4.0 & 10.0 \inc{6.0}& 80.0 & 76.0 \inc{6.0}\\
open\_laptop        & 46.0 & 54.0\inc{8.0}  & 84.0 & 84.0 \\
move\_can\_pot      & 2.0 & 4.0 \inc{2.0}& 56.0 & 64.0 \inc{8.0}\\
rotate\_qrcode      & 16.0 & 20.0 \inc{4.0} & 66.0 & 68.0 \inc{2.0}\\
grab\_roller        & 78.0 & 76.0 \dec{2.0} & 84.0 & 96.0 \inc{12.0}\\
\midrule
average             & 31.5 & 33.5 \inc{2.0} & 69.8 & 73.5 \inc{3.7} \\
\bottomrule
\end{tabular}
\end{table}

\subsection{Experiments on LIBERO and LIBERO-Plus}
As showed in Table~\ref{tab:libero-comparison}, we also evaluate on LIBERO using $\pi_0$ and $\pi_{0.5}$ and observe
a slight performance drop when training with augmented
data: $\pi_0$ decreases by an average of 0.2 and $\pi_{0.5}$ by 0.5 across four task suites. This slightly drop is cause by the evaluation environment of LIBERO is almost identical to original training data, making aggressive augmentation introduce  distributional disturbances. What 's more, with the performance of $\pi_0$ and $\pi_{0.5}$  approach full scores, it can hardly evaluate improvement.

\begin{table}[h!]
\centering
\caption{Performance of $\pi_0$ (mixture strategy) and $\pi_{0.5} $(replacement strategy) on LIBERO.}
\label{tab:libero-comparison}
\setlength{\tabcolsep}{3mm}
\begin{tabular}{l|ll|ll}
\toprule
\multirow{2}{*}{\textbf{Task Name}} & \multicolumn{2}{c|}{\textbf{$\pi_0$}} & \multicolumn{2}{c}{\textbf{$\pi_{0.5}$}}\\
~ & Ori. & Aug. & Ori. & Aug.\\
\midrule
libero spatial     &96.8 & 96.4  & 98.8 & 97.8  \\
libero object & 98.8 & 98.5 & 98.2 & 98.4  \\
libero goal& 95.8 & 96.1 & 98.0 & 98.2 \\
libero 10 & 85.2 & 84.7& 92.4 & 91.0 \\
average  &94.1 & 93.9  & 96.8 & 96.3 \\
\bottomrule
\end{tabular}
\end{table}\par

\textbf{Comparison of mixture and replacement strategies on LIBERO-Plus.} In Table~\ref{tab:openpi-comparison_strategy}, by comparing the use of mixture and replacement on both $\pi_0$ and $\pi_{0.5}$, it is proved that using both strategy can help to boost models' performance on evaluation benchmark. However, whether to choose the mixture strategy or the replacement strategy depend on models' features. For example, using the mixture strategy on $\pi_0$ can surpass using replacement strategy.  

\begin{table}[h!]
\centering
\caption{Performance (\%) comparison on spatial suite from LIBERO-Plus using $\pi_0$ and $\pi_{0.5}$, ``Rep'' means using the Replacement strategy, ``Mix'' means using the mixture strategy}
\label{tab:openpi-comparison_strategy}
\setlength{\tabcolsep}{3mm}
\begin{tabular}{l|ccc|ccc}
\toprule
\multirow{2}{*}{\textbf{Task Name}} & \multicolumn{3}{c|}{\textbf{$\pi_0$}} & \multicolumn{3}{c}{\textbf{$\pi_{0.5}$}}\\
~ & Ori. & Aug. (Rep) & Aug. (Mix)& Ori. & Aug. (Rep)&Aug. (Mix)\\
\midrule
light conditions     & 75.0 & 75.7 & 78.7 & 94.5& 97.9& 99.3 \\
objects layout & 69.6 & 66.2& 86.2 & 97.9& 97.4 & 97.1  \\
background textures & 81.1 & 83.4& 87.6 & 95.7&  95.3& 95.3\\
sensor noise & 19.9 & 16.2& 18.2 &91.2& 93.4 & 93.4 \\
language instructions  &37.9 & 43.2&55.9 & 90.0 & 90.0& 89.1 \\
robot initial states     & 10.3 & 8.1& 6.3 & 82.6&84.9  & 83.2 \\
camera view points     & 21.3 & 17.4& 15.2  & 79.3& 79.5 & 79.3  \\
\midrule
average             & 42.7 &43.3 &47.8  & 89.8& 90.8 & 90.2 \\
\bottomrule
\end{tabular}
\end{table}\par

\subsection{ Computation analysis}
\begin{figure}[H]
\centering
\includegraphics[width=0.5\linewidth]{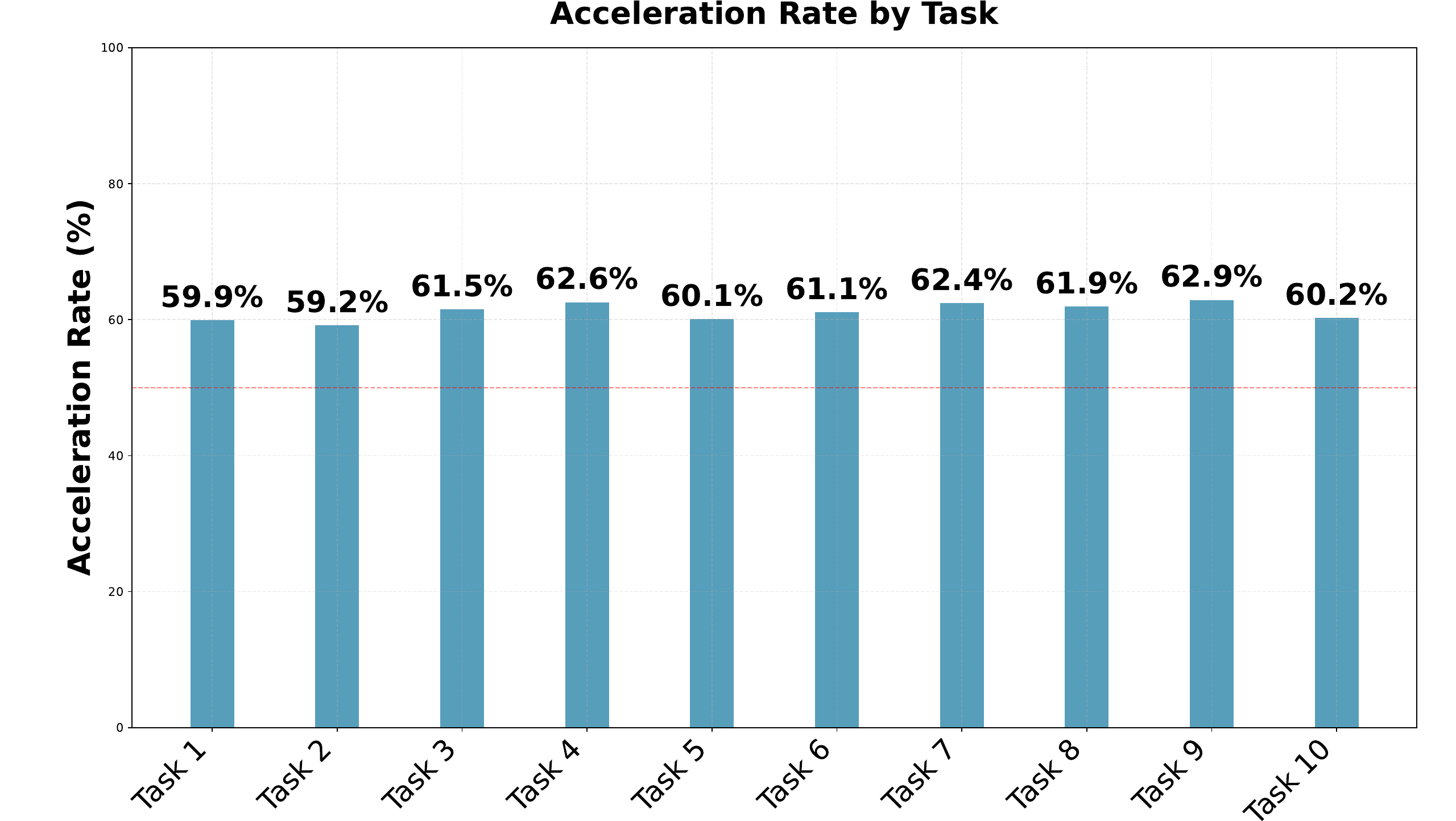}
\caption{Acceleration rates across 10 Robotwin 2.0 tasks. Our method reduces runtime by over 60\% on average. Task 1-10: beat block hammer, adjust bottle, handover block, haing mug, pick dual bottles, place a2b right, place burger fries, place dual shoes, stack blocks two, pick diverse bottles.}
\label{fig:acc}
\end{figure}

During the full procedure for transferring video, the computation costs is computed based on setting the pixels of input video to 960*720 (width, height). We use one A800 80GB GPU for generating, the average time saving percentage (Acceleration Rate) is showed in Figure~\ref{fig:acc}. The average time on the selected ten tasks to augment one video (episode) is 560 seconds on one GPU. Augmenting the full dataset of 50 videos requires 7.7 GPU hours. Training the RDT-1B model on 50 episodes requires 56 GPU hours, augmentation adds only 13.8\% overhead.

\subsection{Justification on hyperparameters}
The choice of k=0.4 and a=8 when transferring videos is based on a trade-off between efficiency and visual–geometric fidelity, validated through ablations on representative task: Adjust Bottle.
We vary k from 0.2 to 0.8 while keeping a=8 constant. As shown in Table~\ref{tab:ablation_k}, increasing k generally leads to a slight degradation in geometric fidelity (e.g., RMSE and Abs.Rel increase), while time cost decreases marginally. However, the overall performance remains stable.\par
\begin{table}[h]
\centering
\caption{Varying k (with a=8)}
\label{tab:ablation_k}
\begin{tabular}{cccccc}
\toprule
$k$ & RMSE & Abs.Rel & Sq.Rel & sim & time cost \\
\midrule
0.2 & 0.26 & 0.14 & 0.06 & 25.04 & 467s \\
0.4 & 0.28 & 0.16 & 0.07 & 26.37 & 441s \\
0.6 & 0.24 & 0.13 & 0.04 & 25.17 & 446s \\
0.8 & 0.32 & 0.18 & 0.08 & 25.16 & 425s \\
\bottomrule
\end{tabular}
\end{table}

We then fix k=0.4 and vary a from 4 to 12. As reported in Table~\ref{tab:ablation_a}, the configuration a=8 achieves the best balance: it maintains high semantic alignment with stable geometry (lowest Sq.Rel and competitive RMSE/Abs.Rel), while achieving the lowest time cost (441 s). Smaller values of a lead to higher time costs without significant accuracy gains, while larger values of a degrade geometric fidelity.\par
\begin{table}[h]
\centering
\caption{Varying a (with k=0.4)}
\label{tab:ablation_a}
\begin{tabular}{cccccc}
\toprule
$a$ & RMSE & Abs.Rel & Sq.Rel & sim & time cost \\
\midrule
4 & 0.26 & 0.16 & 0.06 & 25.47 & 501s \\
6 & 0.24 & 0.15 & 0.06 & 25.14 & 446s \\
8 & 0.28 & 0.16 & 0.07 & 26.37 & 441s \\
12 & 0.26 & 0.14 & 0.05 & 25.36 & 395s \\
\bottomrule
\end{tabular}
\end{table}
We further validate robustness on real-world tasks using the same parameters. The results in Table~\ref{tab:real_world} show that the chosen configuration generalizes well across both simulated and real-world scenarios.\par
\begin{table}[h]
\centering
\caption{Real-world Tasks Validation}
\label{tab:real_world}

\begin{tabular}{lcccc}
\toprule
Task & RMSE & Abs.Rel & Sq.Rel & sim  \\
\midrule
slot pen into holder & 0.156 & 0.150 & 0.02 & 27.04  \\
stack pipe & 0.20 & 0.14 & 0.03 & 26.8 \\
\bottomrule
\end{tabular}
\end{table}

\section{Implementation Details}

\subsection{Using details of cosmos-transfer2.5}
We use original video as input, depth video got from depth anything and prompt generated by our method as control condition. The input pixels is set to be 960*720 (height, width). The diffusion steps totally is set to 35. The number of frames per chunk is set to be 93.

\subsection{LLM Rewriter} \label{sec:appendix_llm}
The designed prompt we using to ask Qwen3-8B rewrite the conditional prompt is as follows where ``{video\_description}'' means video caption, ``{task\_instruction}'' means instruction got from original dataset.

\begin{promptbox}
 \textbf{\blue{<user>:}} You are given a video caption describing a robot manipulation scene and a brief origin video description. Your task is to generate a new video caption for a text-to-video generation model. The new captions should:
 
 * Only change the table surface or objects surface like changing material  to create  differences.
 
 * Output style is expected to look like the following strictly: The video is a demonstration of robotic manipulation, likely in a laboratory or industrial setting. It features a single robotic arm interacting with a plastic bottle. The setting is a room with a polished stainless steel countertop, which reflects overhead lights and provides a sterile, metallic backdrop for the activity. The robotic arm, marked `AGILE X', is positioned above the bottle, which is filled with a dark liquid. At the beginning, the bottle is standing upright on the counter. The robotic arm approaches the bottle, its gripper maneuvering with precision as it positions itself. The arm's gripper then grasps the bottle firmly by its neck. As the arm lifts the bottle smoothly, the liquid inside sways gently. The entire process highlights the precision and control of the robotic arm. The camera remains static throughout, focusing on the interaction between the robotic arm and the bottle, allowing viewers to observe the detailed movements involved in the task.

 * The central focus of the caption should be on a table's surface. The table should be made of wood, stainless steel, or marble.

 * Output is expected to be brief and easy enough for diffusion model to understand.
 
 * The final output should contain only the new caption with no additional commentary or explanation. The video caption is: \textbf{\{video\_description\}}
 
 * The brief origin video description from instruction is: \textbf{\{task\_instruction\}}, make sure that the output content has the same meaning and object name as the given brief description.

 \vspace{4mm}

 \textbf{\green{<example output 1>:}} The video is a demonstration of robotic manipulation, likely in a laboratory or industrial setting. It features a single robotic arm interacting with a green plastic bottle. The setting is a room with a polished wooden table, which provides a warm, natural backdrop for the activity. The robotic arm, marked 'AGILE', is positioned above the bottle. At the beginning, the bottle is standing upright on the table. The robotic arm approaches the bottle, its gripper maneuvering with precision as it positions itself. The arm's gripper then grasps the bottle firmly by its neck. As the arm lifts the bottle smoothly, the process highlights the precision and control of the robotic arm. The camera remains static throughout, focusing on the interaction between the robotic arm and the bottle, allowing viewers to observe the detailed movements involved in the task.

 \textbf{\green{<example output 2>:}} The video is a demonstration of robotic manipulation, likely in a laboratory or industrial setting. It features a single robotic arm interacting with a metal kitchen pot. The setting is a room with a polished wooden countertop, which contrasts with the metallic elements and provides a warm, natural backdrop for the activity. The robotic arm, marked 'AGILE X', is positioned above the pot, which is empty and neatly placed. At the beginning, the kitchen pot is resting on the counter. The robotic arm approaches the pot, its gripper maneuvering with precision as it positions itself. The arm's gripper then grasps the pot firmly by its handles. As the arm lifts the pot smoothly, the action demonstrates meticulous control. The entire process showcases the precision and dexterity of the robotic arm. The camera remains static throughout, focusing on the interaction between the robotic arm and the pot, allowing viewers to observe the detailed movements involved in the task.

 \textbf{\green{<example output 3>:}} The video is set in an experimental lab, featuring a robotic arm labeled 'AGILE'. The arm hovers above a white sneaker on a blue mat, which rests on an elegant wooden table. Surrounding the sneaker, various items such as a trophy, a can, and a small bell are visible. The robotic arm approaches the sneaker, implying an interaction, possibly for cleaning or maintenance.

 \textbf{\green{<example output 4>:}} The video is a demonstration of robotic manipulation, likely in a laboratory or industrial setting. It features a single robotic arm interacting with a slanted rectangular soap. The setting is a room with a polished marble countertop, which beautifully reflects overhead lights and provides an elegant, smooth backdrop for the activity. The robotic arm, marked 'AGILE X', is positioned above a light wood cabinet. At the beginning, the cabinet drawer is closed. The robotic arm approaches, its gripper maneuvering with precision as it opens the drawer. The arm's gripper then gently grasps the soap, noted for its slight groove pattern, and places it inside the open drawer. The entire process highlights the precision and control of the robotic arm. The camera remains static throughout, focusing on the interaction between the robotic arm, the soap, and the drawer, allowing viewers to observe the detailed movements involved in the task.
\end{promptbox}

\subsection{Velocity Caching Strategy}

Algorithm~\ref{alg:minimal_vpcache} provides pseudocode for our velocity caching mechanism. In practice, we set $k=0.4$, $\alpha=8$, and $m=3$ based on empirical tuning.

\begin{algorithm}[h]
\caption{Three-Stage Velocity  Caching}
\label{alg:minimal_vpcache}
\begin{algorithmic}[1]
\REQUIRE Total steps $N$, stable threshold $k$, cache interval $\alpha$, adjustment steps $m$
\FOR{$t = 0 \to N-1$}
\STATE Determine current phase (Initial, Stable, Adjustment)
\IF{Stable Phase and $t \mod \alpha \neq 0$}
\STATE $v_t \gets v_{\text{cache}}$ \hfill \COMMENT{Reuse previous value}
\ELSE
\STATE $v_t \gets v_\theta(x_t, t)$ \hfill \COMMENT{Recompute}
\STATE Update $v_{\text{cache}} \gets v_t$
\ENDIF
\STATE $x_{t+1} \gets x_t + \Delta t \cdot v_t$
\ENDFOR
\end{algorithmic}
\end{algorithm}

\subsection{Real-World Experiments}\label{appdix:real_world}

\noindent \textbf{Experimental setup.}
We deploy our method on an AgileX Piper manipulator equipped with a wrist-mounted camera for visual observation. Data collection is performed via teleoperation using a master-follower arm configuration. We evaluate on two representative manipulation tasks: (1) \textit{Stack Tape}: ``stacking a brown tape roll on top of another brown tape''; and (2) \textit{Slot Pen}: ``inserting a yellow pen into a holder.'' For each task, we collect approximately 80 teleoperated demonstrations (80 episodes for Stack Tape, 77 episodes for Slot Pen). Using our proposed video augmentation pipeline, we generate an equal number of augmented trajectories at a 1:1 ratio, resulting in 160 and 154 total training episodes respectively.

\noindent \textbf{Distribution shift settings.}
To systematically evaluate generalization capability, we design three test conditions with increasing levels of distribution shift:

\noindent \textit{(1) In-Distribution (I.D.).} Target objects (tape or pen) are randomly displaced within a 5cm range from training positions, using the original white table background.

\noindent \textit{(2) Position Shift (OOD).} Target objects are displaced beyond the 5cm training range, introducing spatial perturbations while maintaining the standard white background.

\noindent \textit{(3) Background Shift (OOD).} Objects remain within the 5cm training range, but the table background is changed from white to black to introduce a visual domain shift.
These three scenarios are visualized in Figure~\ref{fig:distribution_shifts}.

\noindent \textbf{Evaluation protocol.}
Each task is evaluated over 10 trials per test condition. A trial is considered successful if the robot completes the full manipulation sequence: for Stack Tape, the robot must grasp and place the tape stably on top of the target tape; for Slot Pen, the pen must be fully inserted into the holder slot. We compare models fine-tuned on original data (baseline) against models fine-tuned on augmented data (ours) using the official OpenPi framework with default full-parameter fine-tuning settings on an NVIDIA H200 GPU.

\section{Training configurations}
We provide detailed training configurations below:

\textbf{Pi0 for LIBERO and LIBERO-Plus in experiment~\ref{exp:libero}}   followed the standard full fine-tuning  configurations provided by the OpenPi repository, which  was trained for 30,000 steps with a global batch size of 32 .

\textbf{Pi0.5 for LIBERO and LIBERO-Plus in experiment~\ref{exp:libero} } followed  the standard full fine-tuning configurations provided by the OpenPi repository, which was trained for 30,000 steps with a global batch size of 256 .

\textbf{RDT for RoboTwin 2.0 multi-tasks training in experiment~\ref{exp:robotwin}} followed the standard training configurations provided by the RoboTwin 2.0 repository, which was trained for 100,000 steps with a batch size of 16 per GPU on 8 GPUs.

\textbf{RDT for RoboTwin 2.0 single-task training in experiment~\ref{exp:robotwin}} followed the standard training configurations provided by the RoboTwin 2.0 repository, which was trained for 10,000 steps with a batch size of 16 per GPU on 4 GPUs.

\textbf{Pi0 for RoboTwin 2.0 single-task training in experiment~\ref{exp:robotwin}} followed the standard training configurations provided by the RoboTwin 2.0 repository, which was performed for 30,000 steps using the global batch size of 32.

\textbf{ACT for RoboTwin 2.0 single-task training in experiment~\ref{exp:robotwin}} followed the standard training configurations provided by the RoboTwin 2.0 repository, which was trained under a unified setup with a chunk size of 50, batch size of 8, and single-GPU training for 6,000 epochs.

\section{Visualization of Augmented Videos}

\subsection{Augmented Video \textit{vs.} Original Video}
\begin{figure}[H]
    \centering
    \includegraphics[width=0.9\linewidth]{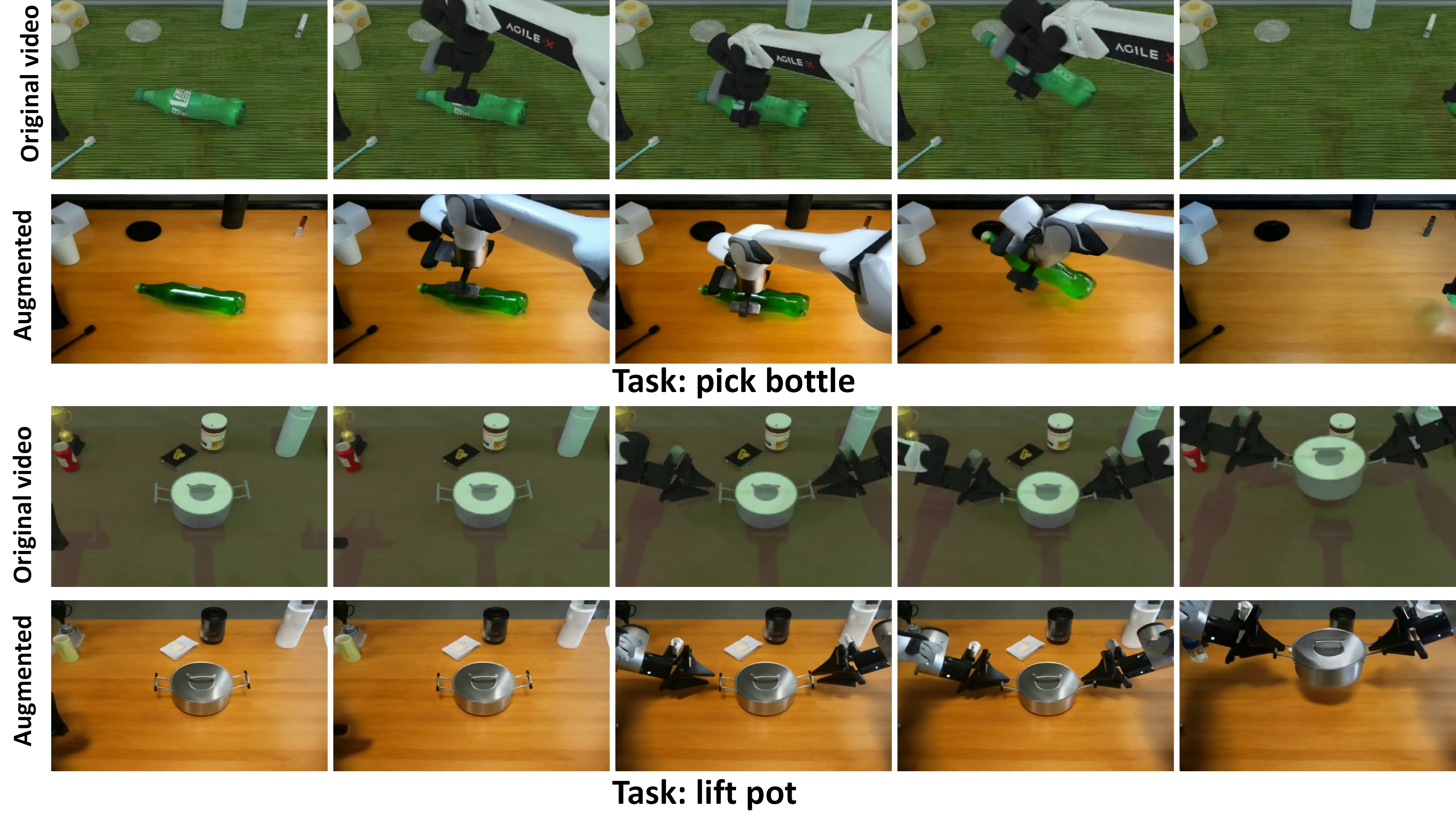}
    \caption{Visualizations of augmented videos and origin videos from Robotwin 2.0 }
    \label{fig:visualize1}
\end{figure}

In Figure~\ref{fig:visualize1} , we visualize the transferred videos along with its original videos got from Robotwin 2.0's domain randomized tasks including lift pot and pick bottle.

\par
\begin{figure}[H]
    \centering
    \includegraphics[width=0.9\linewidth]{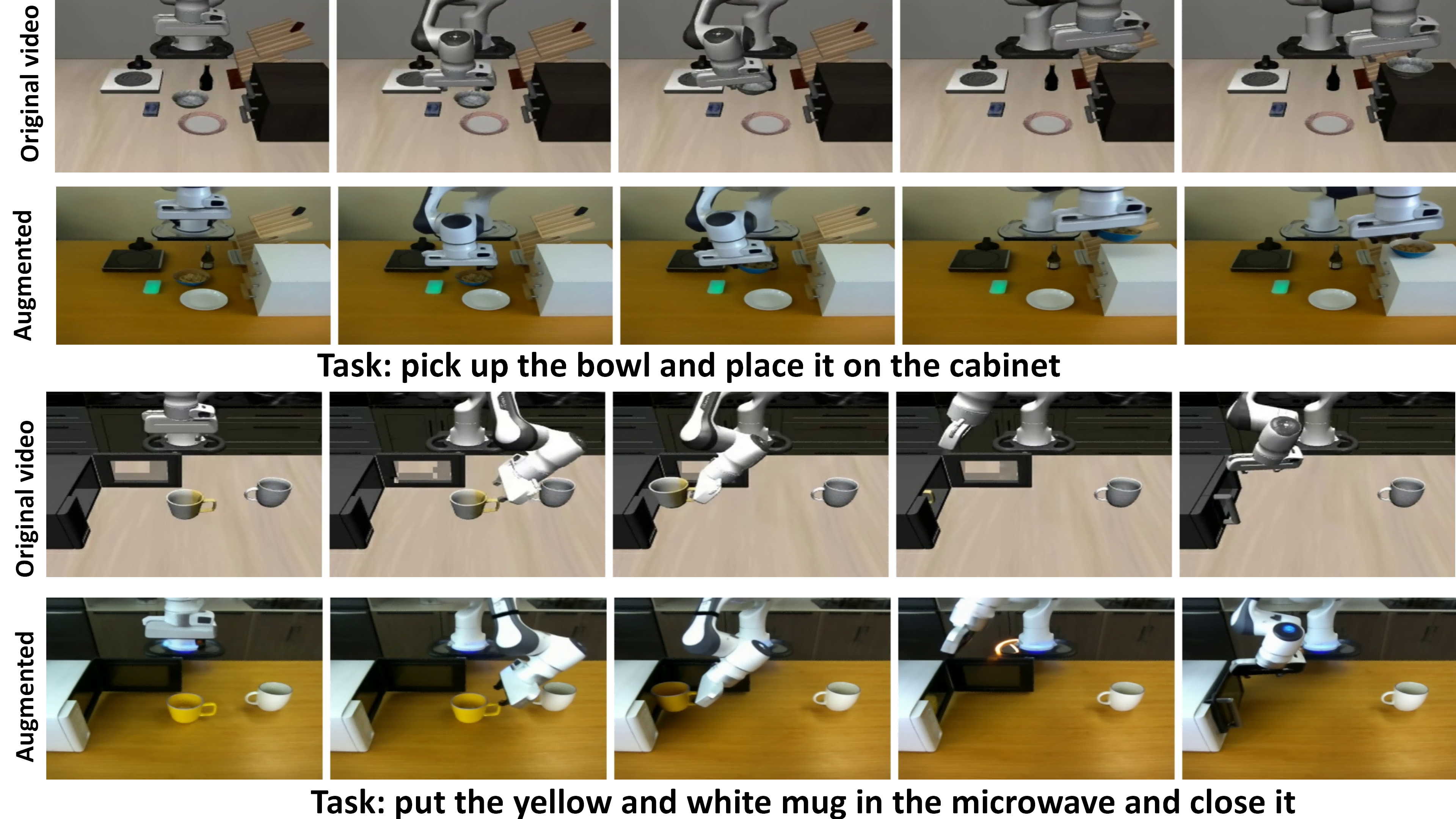}
    \caption{Visualizations of augmented videos and origin videos from LIBERO}
    \label{fig:visualize2}
\end{figure}

In Figure~\ref{fig:visualize2}, we visualize the transferred videos along with its original videos got from LIBERO official dataset.

\par

\begin{figure}[H]
    \centering
    \includegraphics[width=0.9\linewidth]{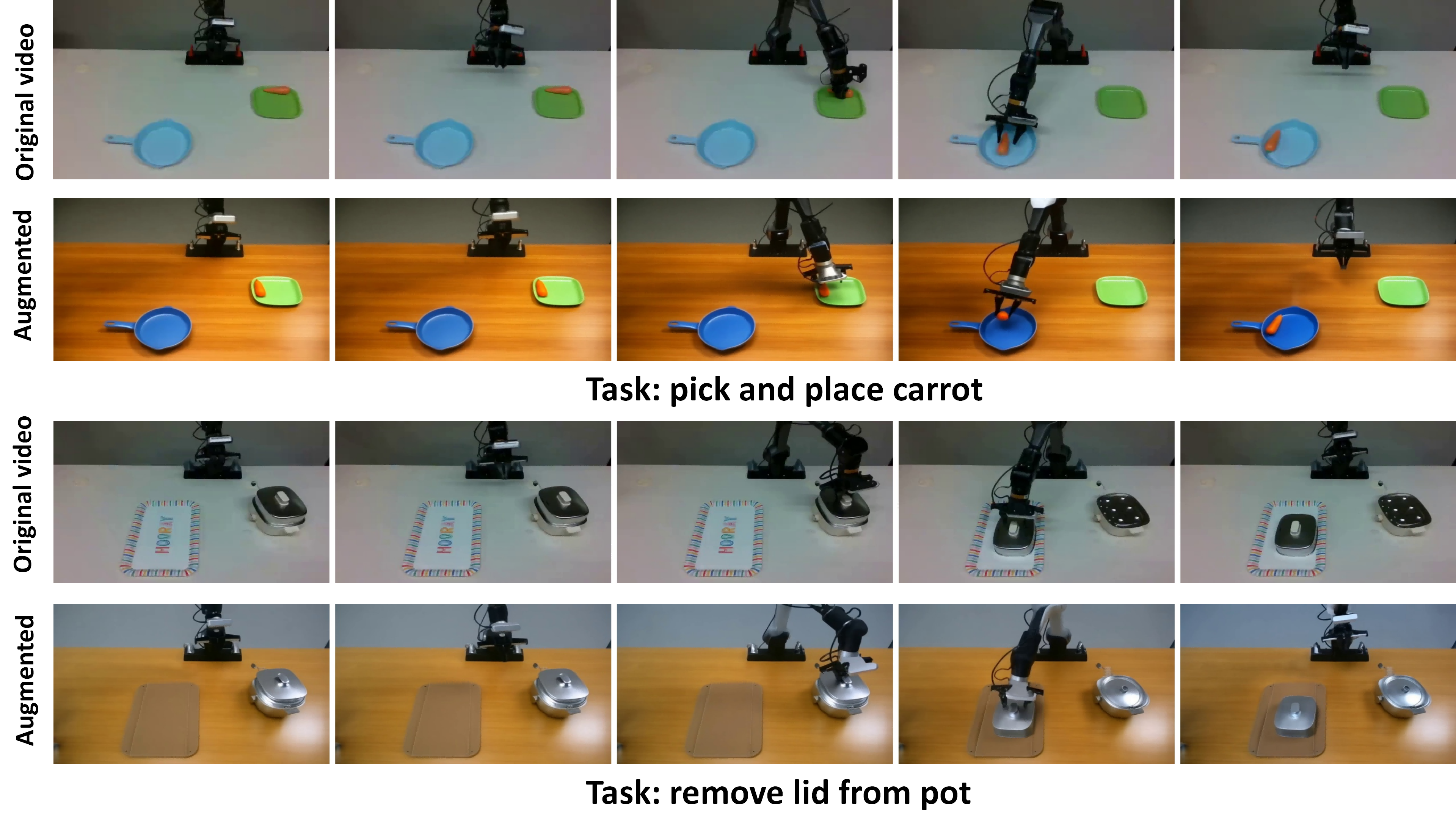}
    \caption{Visualizations of augmented videos and origin videos from real robot experiment}
    \label{fig:visualize3}
\end{figure}

In Figure~\ref{fig:visualize3}, we visualize the transferred videos along with its original videos got from real world recordings including tasks ``pick and place carrot'' and ``remove lid from pot''.

\begin{figure}[H]
    \centering
    \includegraphics[width=0.9\linewidth]{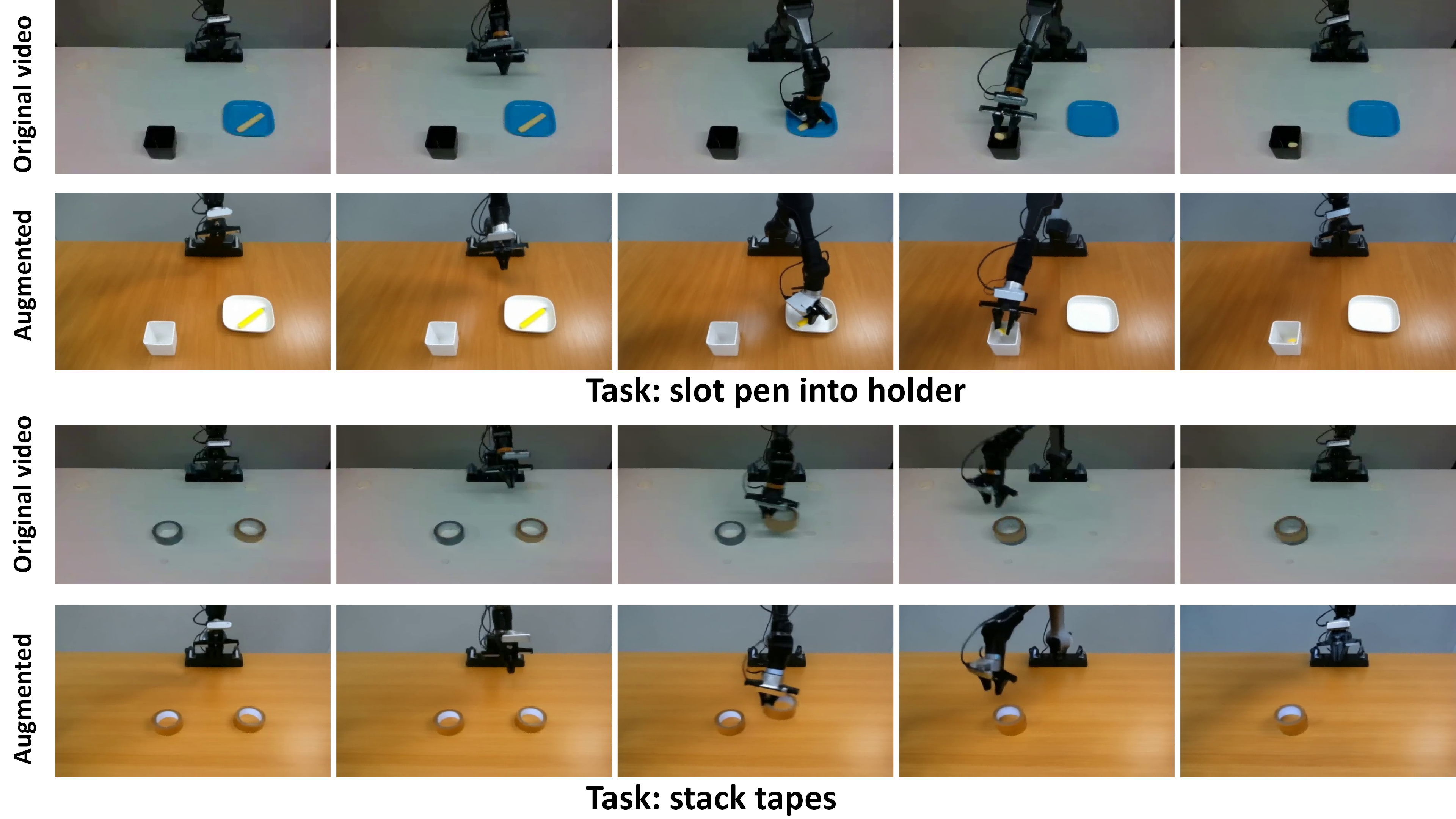}
    \caption{Visualizations of augmented videos and origin videos from real robot experiment}
    \label{fig:visualize4}
\end{figure}

In Figure~\ref{fig:visualize4}, we visualize the transferred videos along with its original videos got from real world recordings including tasks ``slot pen into holder'' and ``stack tapes''.\par

\subsection{Augmenting Videos with or without Velocity Caching}

\begin{figure}[H]
    \centering
    \includegraphics[width=0.9\linewidth]{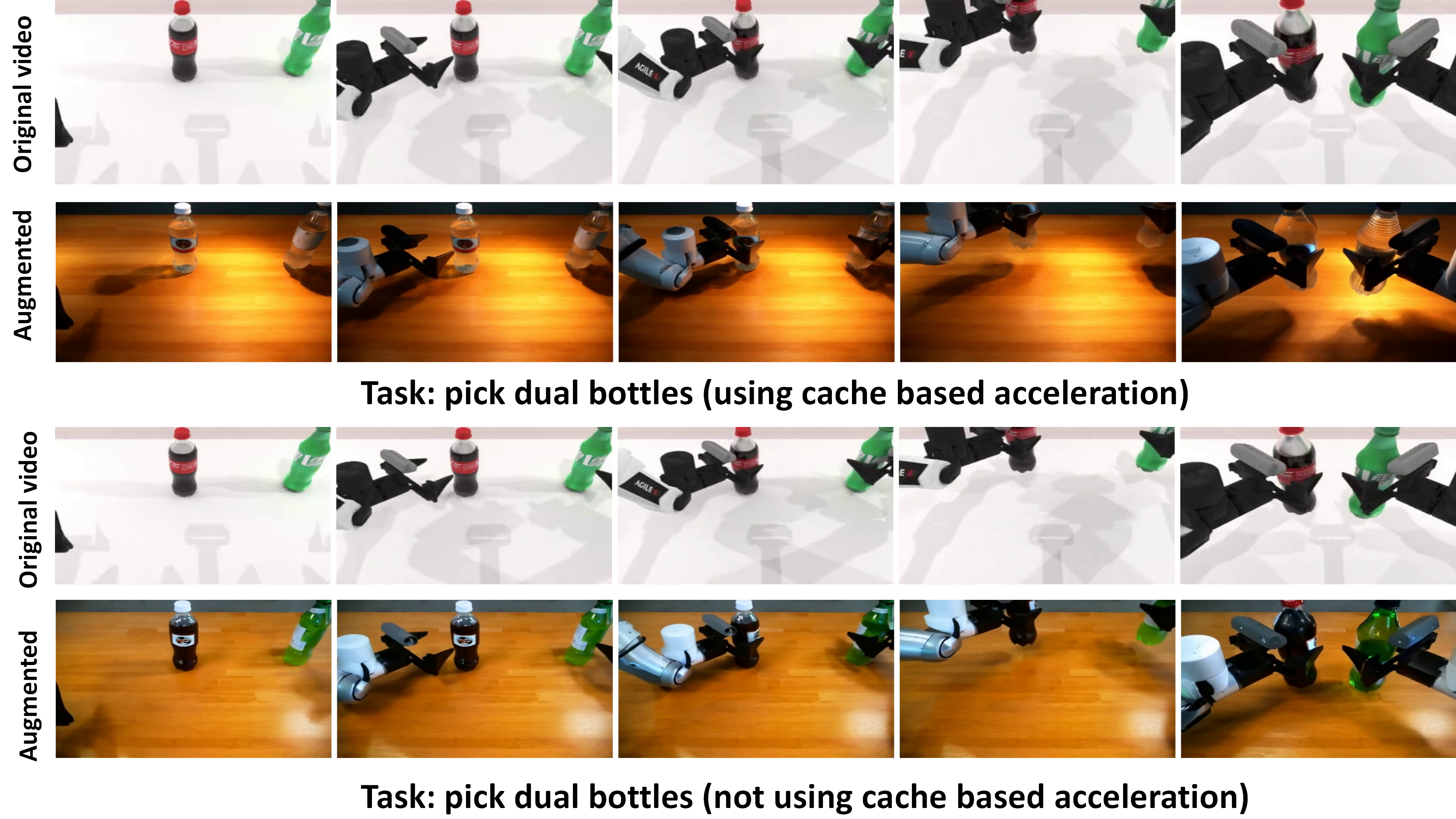}
    \caption{Comparing between using cache based acceleration and not using }
    \label{fig:compareacc1}
\end{figure}

In Figure~\ref{fig:compareacc1}, we visualize the transferred videos along with its original videos got from Robotwin 2.0's domain clean tasks and compare between using acceleration and not using acceleration.

\begin{figure}[H]
    \centering
    \includegraphics[width=0.9\linewidth]{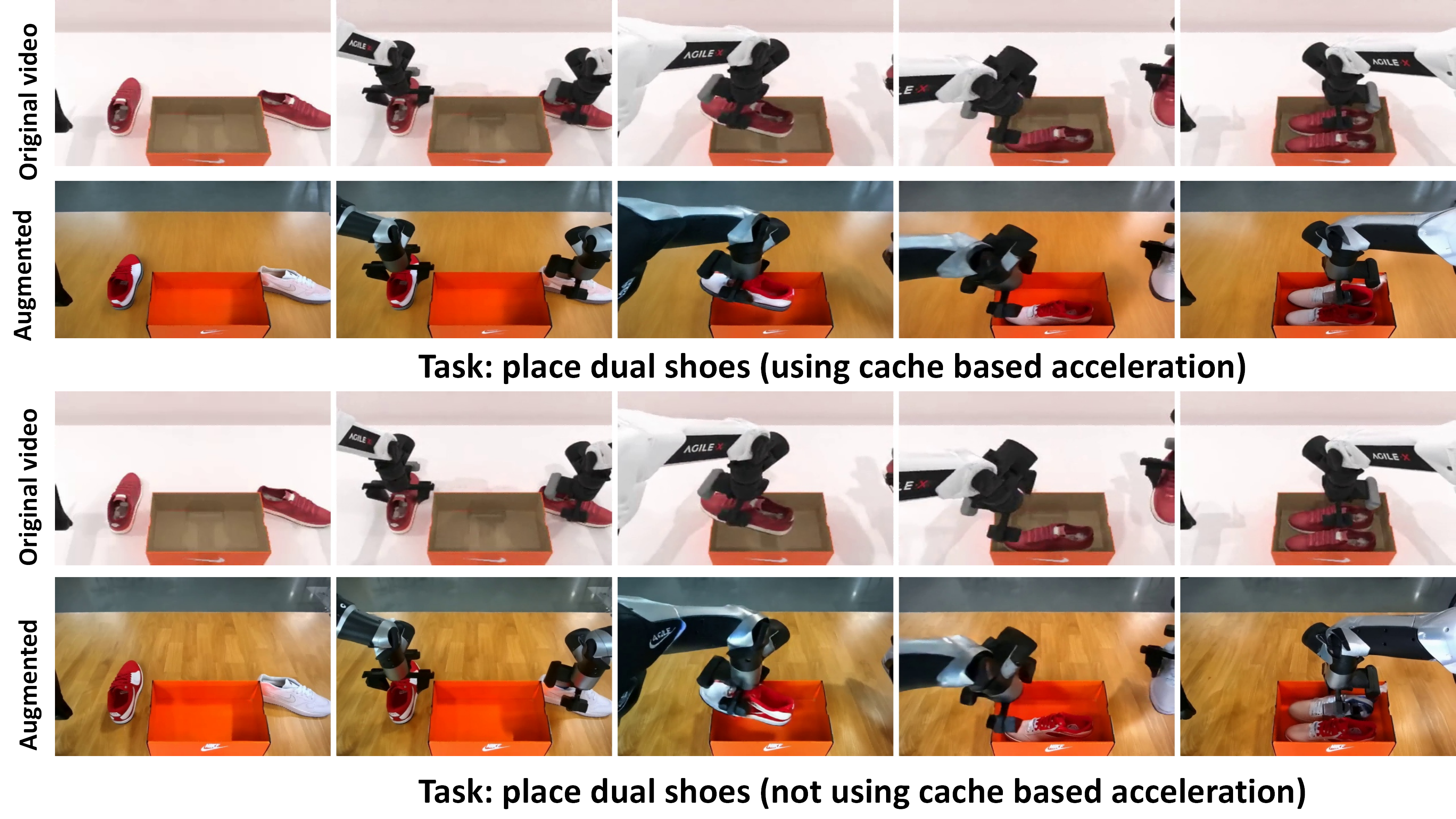}
    \caption{Comparing between using cache based acceleration and not using}
    \label{fig:compareacc2}
\end{figure}

In Figure~\ref{fig:compareacc2}, we visualize the transferred videos along with its original videos got from Robotwin 2.0 's domain clean tasks and compare between using acceleration and not using acceleration.

\end{document}